\documentclass[review]{elsarticle}
\usepackage[margin=2.5cm]{geometry}
\usepackage{lipsum}
\usepackage{amsfonts}
\usepackage{amsmath}
\usepackage{multicol}        
\usepackage{algorithm,algpseudocode}
\usepackage{algorithmicx}
\usepackage{multirow}
\usepackage{subfig}
\usepackage{color}

\journal{Signal Processing: Image Communication}

\begin{document}

\begin{frontmatter}
\title{Block Compressive Sensing of Image and Video with Nonlocal Lagrangian Multiplier and Patch-based Sparse Representation}


\author[Address1]{Trinh Van Chien \corref{mycurrentAddress}}
\author[Address1]{Khanh Quoc Dinh}
\author[Address1]{Byeungwoo Jeon\corref{mycorrespondingauthor}}
\ead{bjeon@skku.edu}
\author[Address2]{Martin Burger}

\address[Address1]{School of Electrical and Computer Engineering, Sungkyunkwan University, Korea}
\address[Address2]{Institute for Computational and Applied Mathematics, University of M\"{u}nster, Germany}
\cortext[mycurrentAddress]{He is now with Communication Systems Division, Department of Electrical Engineering (ISY), Link\"{o}ping University, Sweden.}
\cortext[mycorrespondingauthor]{Corresponding author}

\begin{abstract}
Although block compressive sensing (BCS) makes it tractable to sense large-sized images and video, its recovery performance has yet to be significantly improved because its recovered images or video usually suffer from blurred edges, loss of details, and high-frequency oscillatory artifacts, especially at a low subrate. This paper addresses these problems by designing a modified total variation technique that employs multi-block gradient processing, a denoised Lagrangian multiplier, and patch-based sparse representation. In the case of video, the proposed recovery method is able to exploit both spatial and temporal similarities. Simulation results confirm the improved performance of the proposed method for compressive sensing of images and video in terms of both objective and subjective qualities.
\end{abstract}


\end{frontmatter}


\section{Introduction}

Current video coding techniques, such as HEVC \cite{SullivanOHW12}, are designed to have low-complexity decoders for broadcasting applications; this is based on the assumption that large amounts of resources are available at the encoder. However, many emerging real-time encoding applications, including low-power sensor network applications or surveillance cameras, call for an opposite system design that can work with very limited computing and power resources at the encoder. Distributed video coding (DVC) \cite{hoangvan2012} is an alternate solution for a low-complexity encoder, in which the encoding complexity is substantially reduced by shifting the most computationally-intensive module of motion estimation/motion compensation to the decoder. Nonetheless, other than the encoding process, the processes of image/video acquisition also need to be considered to further reduce the complexity of the encoder \cite{hoangvan2012} because current image/video applications capture large amounts of raw image/video data, most of which are thrown away in the encoding process for achieving highly compressed bitstream. In this context, compressive sensing (CS) has drawn interest since it provides a general signal acquisition framework at a sub-Nyquist sampling rate while still enabling perfect or near-perfect signal reconstruction \cite{donoho2006}. More clearly, a sparse signal that has most entries equal to zero (or nearly zero) can be sub-sampled via linear projection onto sensing bases; this can be reconstructed later by a sophisticated recovery algorithm, which basically seeks its $K$-sparse approximation (i.e., the $K$ largest magnitude coefficients). Consequently, CS leads to simultaneous signal acquisition and compression to form an extremely simple encoder. Despite its simplicity, its recovery performance is heavily dependent on the recovery algorithm, in which some of the important factors are properly designing the sparsifying transforms and deploying appropriate denoising tools.

Although many CS recovery algorithms have been developed, including NESTA (Nesterov's algorithm) \cite{becker2011}, gradient projection for sparse reconstruction (GPSR) \cite{figueiredo2007}, Bayesian compressive sensing \cite{ji2008,he2009,he2010}, smooth projected Landweber (SPL) \cite{gan2007}, and total variation (TV)-based algorithms \cite{li2013,zhang2010,xu2012}, their reconstructed quality has yet to be improved much, especially at a low subrate. For better CS recovery, Candes \cite{candes2008} proposed a weighted scheme based on the magnitude of signals to get closer to $\ell_0$ norm, while still using $\ell_1$ norm in the optimization problems. In a similar manner, Asif et al.~\cite{asif2013} adaptively assigned weight values according to the homotopy of signals. As another approach, the authors in \cite{mun2009, van2013, van2014} utilized local smoothing filters, such as Wiener or Gaussian filters, to reduce blocking artifacts and enhance the quality of the recovered images. Despite these improvements, the performances of the aforementioned approaches are still far from satisfactory because much of the useful prior information of the image/video signals (e.g., the non-local statistics) was not taken into full account.

More recent investigations have sought to design a sparsifying transform to sparsify the image/video signal to the greatest degree because the CS recovery performance can be closer to that of sampling at the full Nyquist rate if the corresponding transform signal is sufficiently sparse \cite{donoho2006}. The direct usage of predetermined transform bases, such as the discrete wavelet transform (DWT) \cite{he2009, mun2009}, discrete cosine transform (DCT) \cite{he2010, mun2009}, or gradient transform \cite{li2013, zhang2010, xu2012, van2014}, is appealing due to their low complexity. However, predetermined transform bases cannot produce sufficient sparsity (i.e., the number of zero or close-to-zero coefficients is limited) for the signal of interest, thereby limiting their recovery performance. Because image and video signals are rich in nonlocal similarities (i.e., a pixel can be similar to other pixels that are not located close to it), usage of those nonlocal similarities \cite{buades2005} can generate a higher sparsity level to achieve better recovery performance; this is known as a patch-based sparse representation approach \cite{elad2006}. Note that this approach originally showed much success in image denoising \cite{elad2006, dabov2007, dabov2009bm3d, chatterjee2012} and researchers have incorporated this idea into CS frameworks. Xu and Yin \cite{xu2014fast} proposed a fast patch method for whole-image sensing and recovery under a learned dictionary, while Zhang et al. \cite{zhang2012image} took advantage of hybrid sparsifying bases by iteratively applying a gradient transform and a three-dimensional ($3$D) transform \cite{dabov2007}. By using the concept of decomposition, the authors in \cite{canh2016compressive} also used a $3$D transform for cartoon images to enhance the recovery quality. The $3$D transform can be considered a global sparsifying transform because it is used for all patches of the recovered images. Dong et al.~\cite{dong2014compressive}, motivated by the success of data-dependent transforms for patches (referred to as local sparsifying transforms) such as principal component analysis (PCA) or singular-value decomposition (SVD), proposed a method to enhance the sparsity level with the \textit{logdet} function to bring $\ell_1$ norm closer to $\ell_0$ norm, similar to the work of Candes \cite{candes2008}. Metzler et al. \cite{metzler2016denoising} acquired a local sparsifying transform via block matching  \cite{dabov2009bm3d} and demonstrated the effectiveness of applying denoising tools to the CS recovery of the approximate message passing (AMP) method. However, because of the frame sensing that accesses the entire image at once, the work described in \cite{xu2014fast, zhang2012image, dong2014compressive, metzler2016denoising} requires extensive computation and huge amounts of memory for storing the sensing matrix \cite{fowler2012block}; thus, these approaches are not suitable as sensing schemes for real-time encoding applications or large-scale images/video.

Alternatively, block compressive sensing (BCS) has been developed to deal more efficiently with large-sized natural images and video by sensing each block separately using a block sensing matrix with a much smaller size. The compressive sensor can instantly generate the measurement data of each block through its linear projection rather than waiting until the entire image is measured, as is done in frame sensing. The advantages of BCS are discussed in \cite{gan2007, fowler2012block, dadkhah2014block, dinh2016iterative}. However, in BCS, the recovery performance has yet to be substantially improved in comparison to that of frame sensing. To address this problem on the sensing side, a Gaussian regression model between the coordinates of pixels and their gray levels can be used to achieve better performance compared to traditional Gaussian matrices \cite{han2015novel}. Additionally, Fowler et al.~\cite{fowler2011multiscale} developed an adaptive subrate method (i.e., multi-scale BCS) to exploit the different roles of wavelet bands. On the recovery side, for example, Dinh et al. \cite{dinh2014weighted} designed overlapped recovery with a weighted scheme to reduce the blocking artifacts caused by block recovery. Chen et al. \cite{chen2011compressed} used the Tikhonov regularization and residual image information to enhance the smooth projected Landweber \cite{gan2007}. Furthermore, to enrich the details of recovered images, the K-SVD algorithm \cite{elad2006} was used in \cite{zhang2014image}. By sharing the same idea in \cite{xu2014fast, zhang2012image, dong2014compressive, metzler2016denoising, zhang2014image} where nonlocal similarities are exploited to design the local sparsifying transform, group-based sparse representation (GSR) \cite{zhang2014group} can achieve better recovery performance (in terms of the peak signal-to-noise-ratio (PSNR)) than other algorithms that were previously designed for BCS. However, its recovered images still contain many visual artifacts since the nonlocal searching and collecting patches based on the initial recovered images produced by \cite{chen2011compressed} often have poor quality at low subrates. Consequently, this implies that more efforts are required for improving both the objective and subjective quality.

This paper attempts to improve the recovery performance of the BCS framework by using TV minimization, which is good at preserving edges \cite{li2013}, with multiple techniques consisting of reducing blocking artifacts in the gradient domain, denoising the Lagrangian multipliers, and enhancing the detailed information with patch-based sparse representation. Furthermore, the proposed recovery methods are easily extendible to compressive sensing and encoding problems of video \cite{do2009distributed, mun2011residual, tramel2011video, van2014block, kang2009distributed}. Specifically, our main contributions are summarized as follows.
\begin{itemize}
\item For BCS of images, we propose a method, referred to as multi-block gradient processing, that addresses the blocking artifacts caused by block-by-block independent TV processing during recovery. Furthermore, based on our observation that both image information (e.g., edges and details) and high-frequency artifacts and staircase artifacts are still prevalent in the Lagrangian multiplier of the TV optimization, we propose a method to reduce such artifacts by denoising the Lagrangian multiplier directly with a nonlocal means (NLM) filter. Because the direct application of the NLM filter is not effective in preserving local details with low contrast \cite{buades2005}, we further propose enriching these low-contrast details through an additional refinement process that uses patch-based sparse representation. We propose using both global and local sparsifying transforms because the single usage of either transform limits the effective sparse basis and achievement of a sufficient sparsity level for noisy data. The proposed recovery method demonstrates improvements for BCS of images compared to previous works  \cite{he2009, he2010, mun2009, van2013, chen2011compressed, zhang2014image, zhang2014group}.

\item For BCS of videos, we extend the proposed method to a compressive video sensing problem known as block distributed compressive video sensing (DCVS). An input video sequence is divided into groups of pictures (GOP), each of which consists of one key frame and several non-key frames. These undergo block sensing by a Gaussian sensing matrix. The proposed method first recovers the key frame using the proposed recovery method. Then, for non-key frames, side information is generated by exploiting measurements of the non-key and previously recovered frames in the same GOP. Improved quality of the non-key frames is sought by joint minimization of the sparsifying transforms and side information regularization. Our experimental results demonstrate that the proposed method performs better than existing recovery methods designed for block DCVS, including BCS-SPL using motion compensation (MC-BCS-SPL) \cite{mun2011residual} or BCS-SPL using multi-hypothesis prediction (MH-BCS-SPL) \cite{tramel2011video}.
\end{itemize}
The rest of this paper is organized as follows. Section~\ref{SecII} briefly presents works related to the BCS framework with some discussion. The proposed recovery method for BCS of images is described in Section~\ref{SecIII}, and its extension to the block DCVS model is addressed in Section~\ref{SecIV}. Section~\ref{SecV} evaluates the effectiveness of the proposed methods compared to other state-of-the-art recovery methods. Finally, our conclusions are drawn in Section ~\ref{SecVI}.

\section{Block compressive sensing} \label{SecII}

In the BCS framework, a large-sized image $u$ is first divided into multiple non-overlapping (small) blocks. Let a vector $\bar{u}_k$ of length $n$ denote the $k$th block, which is vectorized by raster scanning. Its $m \times 1$ measurement vector $b_k$ is generated through the following linear projection by a sensing matrix $A_B:$
\begin{equation} \label{Eq1}
b_k = A_B \bar{u}_k
\end{equation}
A ratio $(m/n)$ denotes the subrate (or sub-sampling rate, i.e., the measurement rate). BCS is memory-efficient as it only needs to store a small sensing matrix instead of a full one corresponding to the whole image size. In this sense, block sampling is more suitable for low-complexity applications. 

The CS recovery performance heavily depends on the mutual coherence $\chi$ of the sensing matrix $A_B$ which is computed as \cite{eldar2012compressed}:
\begin{equation} \label{Eq2}
\chi = \underset{1 \leq i <j \leq n}{\textrm{max}} \quad \frac{|\langle\,a_i,a_j\rangle|}{ \| a_i \|_2  \| a_j \|_2}
\end{equation}
Here, $a_i$  and $a_j$ are any two arbitrary columns of $A_B$; $\langle\,,\rangle$  denotes the inner product of two vectors. According to the Welch bounds \cite{eldar2012compressed}, $\chi$ is limited in the range of $[ \sqrt{(n-m)/ m(n-1)}, 1]$. Additionally, at a low subrate (i.e., $m \ll n$), it can be approximated as
\begin{equation} \label{Eq3}
\chi \in \left[\frac{1}{\sqrt{m}},1\right]
\end{equation}
Note that a low mutual coherence is preferred for CS, and that the lower bound of \eqref{Eq3} is inversely proportional to $\sqrt{m}$. A low mutual coherence is harder to achieve with a small block size, although it is attractive in terms of memory requirements. This explains the limited recovery quality of BCS with a small block size, despite the great amount of research that has been conducted on this topic \cite{he2009, he2010, mun2009, dinh2014weighted, chen2011compressed, zhang2014image, dinh2014measurement, kulkarni2016reconnet}.  

The work in \cite{he2009, he2010} sought to obtain structured sparsity of signals in the Bayesian framework with a Markov-chain Monte-Carlo \cite{he2009} or variational Bayesian framework \cite{he2010}. However, for practical image sizes, these approaches demand impractical computational complexity; thus, the search must be terminated early, and the recovered images suffer from high-frequency oscillatory artifacts \cite{zhang2012image}. The recoveries \cite{mun2009, chen2011compressed} are much faster than those previous ones, but they are limited by their predetermined transforms. Specifically, non-iterative recovery was demonstrated in \cite{kulkarni2016reconnet} with a convolutional neural network being used for the measurement data. The work in \cite{zhang2014image, zhang2014group} uses a single patch-based sparse representation; therefore, its reconstructed quality is still not satisfactory. The initial images used for adaptive learning of sparsity contain a large amount of noise and artifacts; thus, using only one of them limits the definition of the effective sparse basis and makes it difficult to achieve a sufficient sparsity level for noisy data. These observations motivated us to design a new recovery method for high recovery quality, as discussed in the next section.

\section{Proposed recovery for block compressive sensing of image} \label{SecIII}

In this section, we design a recovery method for BCS of images. For this, we modify the TV-based recovery method \cite{li2013} for BCS by introducing a multi-block gradient process to reduce blocking artifacts and directly denoise the nonlocal Lagrangian multiplier to mitigate the artifacts generated by the TV-based methods. Also, due to the limitations of the NLM filter in preserving the image texture, a patch-based sparse representation is also designed to enrich the local details of recovered images.

\subsection{CS recovery for BCS framework}
\begin{figure}
	\centerline{%
		\setlength\tabcolsep{1.5pt} 
		\begin{tabular}{c c} 
			\subfloat[SPLDDWT \cite{mun2009} ]{\label{fig1a}\includegraphics[width= 4cm, height = 4cm]{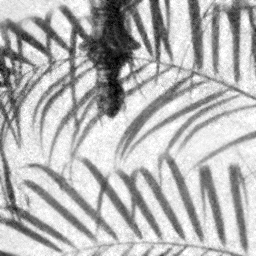}} &
			\subfloat[TV \cite{li2013}]{\label{fig1b}\includegraphics[width= 4cm, height = 4cm]{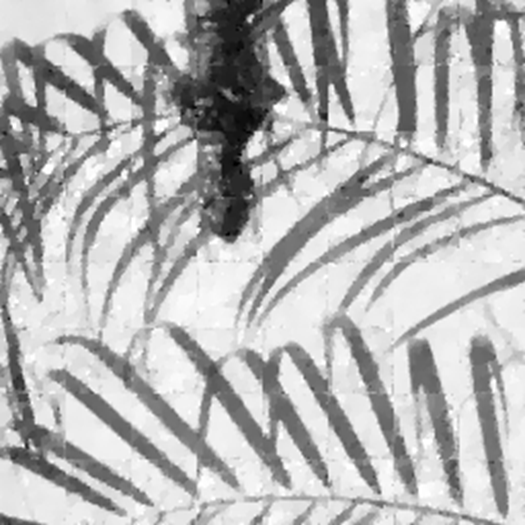}}
		\end{tabular}}
		\caption{BCS recovered image of the test image Leaves (subrate $0.2$, block size $32 \times 32$)} \label{fig1}
	\end{figure}

The first recovery method for BCS was proposed by L. Gan \cite{gan2007}, who incorporated a Wiener filter and Landweber iteration with the hard thresholding process. S. Mun et al. \cite{mun2009} further improved this idea by adding directional transforms of DCT, dual-tree DWT (DDWT), or contourlet transform (CT). The visual quality of the best method \cite{mun2009}, namely the SPL with the dual-tree wavelet transform (SPLDDWT), is shown in Figure~\ref{fig1a} for the image Leaves. The recovered image suffers from high-frequency oscillatory artifacts \cite{candes2006stable} and has blurred edges. TV has been shown to be effective for frame CS \cite{li2013} in preserving edges and object boundaries in recovered images. As expected, the recovered image by TV, shown in Figure~\ref{fig1b}, looks much sharper than the image recovered via SPLDDWT \cite{mun2009}, although TV is applied to BCS for block-by-block recovery. However, it shows significant blocking artifacts, due to the block-independent TV processing. Motivated by this, we investigate a TV-based recovery scheme for BCS with an emphasis on improving the TV method that is applied to BCS such that it does not suffer from blocking artifacts. Our TV-based BCS recovery method has several implications, as follows.

\subsection{Noise and artifacts reduction}
As mentioned above, independent block-by-block TV processing makes images suffer heavily from blocking artifacts as in Figure~\ref{fig1b}. When TV is computed separately for individual blocks as in \cite{xu2012}, a good de-blocking filter should also be used to mitigate the blocking artifacts. When the BCS scheme \cite{gan2007} is in use, a block diagonal sensing matrix $A$ corresponding to a whole image $u$ of size $\sqrt{N} \times \sqrt{N}$ (assuming it consists of $G$ blocks) and its measurement $b$ are given as
\begin{align} 
A &= \textrm{diag} \left( A_B, \ldots, A_B \right) \label{Eq4}\\ 
b &= \left[ b_1 b_2 \ldots b_G \right] \label{Eq5}
\end{align}

\begin{figure}
	\centerline{%
		\setlength\tabcolsep{1.5pt} 
		\begin{tabular}{c c c c} 	
			\subfloat[]{\label{fig2a} \includegraphics[trim=7cm 14.4cm 7.5cm 6.5cm, clip=true, width=1.4in]{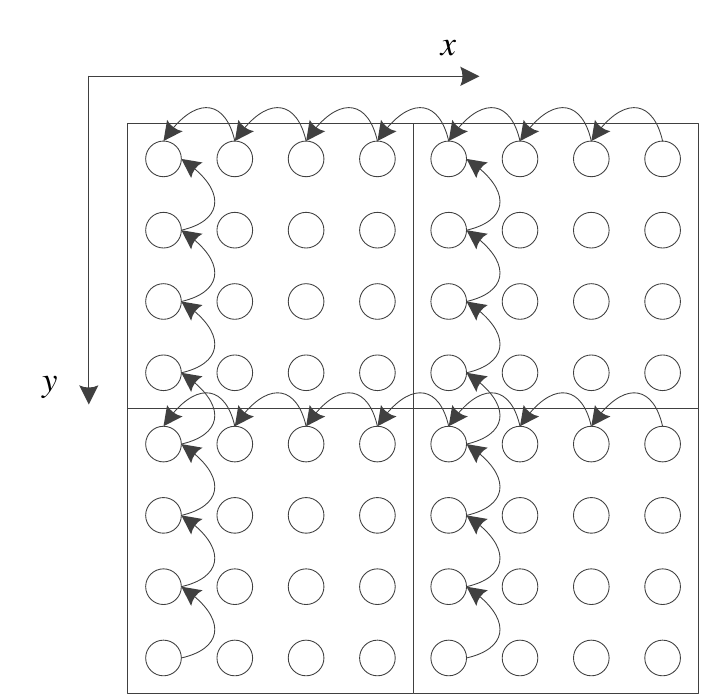}} &
			\subfloat[]{\label{fig2b} \includegraphics[width= 3cm, height = 3cm]{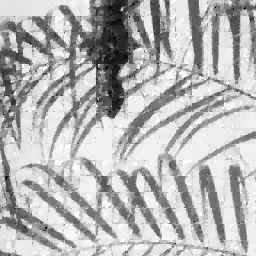}} &
			\subfloat[]{\label{fig2c} \includegraphics[width= 3cm, height = 3cm]{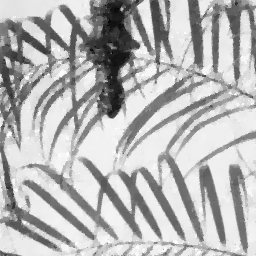}} &
			\subfloat[]{\label{fig2d} \includegraphics[width= 3cm, height =3cm]{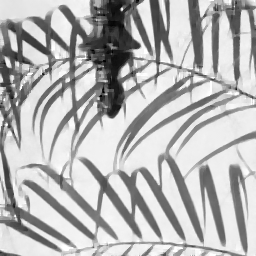}}
		\end{tabular}}
		\caption{Multi-block total variation (MBTV) based on multi-block gradient process. $(a)$ Multi-block gradient process; the recovered images of the test image Leaves at subrate $0.2$,  with a block size of $16 \times 16$, as recovered respectively by $(b)$ TV with block independent gradient process, $(c)$ the proposed MBTV, and $(d)$ the proposed MBTV with nonlocal Lagrangian multiplier denoising.} \label{fig2}
\end{figure}

We design a method, referred to as multi-block total variation (MBTV), based on a multi-block gradient process, as depicted in Figure~\ref{fig2a}. This calculates the gradient for TV over multiple blocks such that the discontinuities at block boundaries can be reduced significantly by minimizing the gradient. Notice that, if this method is applied to all blocks in a recovered image, it is equivalent to a frame-based gradient calculation. The visual quality of the recovered image Leaves, which is illustrated in Figure~\ref{fig2c}, demonstrates that many of the blocking artifacts are reduced compared to the block-by-block TV-based recovery (Figure~\ref{fig2b}). Here, we use a small sensing matrix ($16 \times 16$) to visualize the significant improvements of MBTV. Note that the recovered images usually suffer from more blocking artifacts with a small sensing operator when they are independently recovered block-by-block, as shown in Figure~\ref{fig2b}.

The proposed MBTV-based recovery is described in detail below. The constrained problem of TV-based CS is expressed as 
\begin{equation} \label{Eq6}
\underset{u}{\textrm{min}} \quad TV(u) \qquad \textrm{s.t.} \quad Au =b
\end{equation}
where  $TV(.)$ stands for the TV operator, which can be either isotropic or anisotropic. For isotropic TV, \eqref{Eq6} is converted into an unconstrained problem by the augmented Lagrangian method \cite{li2013, van2014} 
 \begin{equation} \label{Eq7}
\mathcal{L}(w,u) = \left\{ \| w \|_2 - \upsilon^T (Du -w) + \frac{\beta}{2} \|Du -w \|_2^2  - \lambda^T (Au -b) + \frac{\mu}{2} \|Au -b \|_2^2 \right\}
 \end{equation}
Here, $w = Du$ , and $D \in \{ D_x, D_y\}$, where  $D_x$  and $D_y$  denote the horizontal and vertical gradient operators, respectively. $\upsilon$ and $\lambda$ are Lagrangian multipliers,  and  $\beta$ and $\mu$  are positive penalty parameters. The key idea of the augmented Lagrangian method is to seek a saddle point of $\mathcal{L}(w,u)$  that is also the solution of \eqref{Eq6}. At the $(t+1)$th iteration, by acquiring the splitting technique \cite{afonso2010fast}, \eqref{Eq7} is iteratively solved by two so-called sub-problems, $w^{t+1}$ and $u^{t+1}$ , as shown below:
\begin{align} 
w^{t+1} &= \underset{w}{\textrm{argmin}} \left\{ \| w \|_2 - \upsilon^T (Du^t -w) + \frac{\beta}{2} \|Du^t -w \|_2^2 \right\} \label{Eq8} \\
u^{t+1} &= \underset{u}{\textrm{argmin}} \left\{ - \upsilon^T (Du -w^{t+1}) + \frac{\beta}{2} \|Du -w^{t+1} \|_2^2  - \lambda^T (Au -b) + \frac{\mu}{2} \|Au -b \|_2^2 \right\} \label{Eq9}
\end{align}
The solution of \eqref{Eq8} is found by the shrinkage-like formula, where $\odot$  denotes the element-wise product: 
\begin{equation} \label{Eq10}
w^{t+1} = \textrm{max} \left\{ \left\|Du^t - \frac{\upsilon^t}{\beta} \right\|_2 - \frac{1}{\beta}, 0\right\} \odot \frac{Du^t - \upsilon^t / \beta}{ \| Du^t - \upsilon^t / \beta\|_2}
\end{equation}
Because \eqref{Eq9} is a quadratic function, its solution can be achieved by calculating the first derivative of the sub-problem $u$. However, to reduce the computational complexity of the Moore-Penrose inverse \cite{li2013}, we also use gradient descent, as proposed in \cite{li2013, van2014}: 
\begin{equation} \label{Eq11}
u^{t+1} = u^t - \eta  d
\end{equation}
The direction $d$ and the optimized step size $\eta$  in \eqref{Eq11} are calculated by the Barzilai$-$Borwein method \cite{li2013, van2013}: 
\begin{align} 
 & d = \beta D^T \left( Du^t - w^{t + 1} \right) - D^T \upsilon^t + \mu A^T \left( Au^t - b \right) - A^T\lambda^t \label{Eq12}\\
& \eta  = \left\langle {d,d} \right\rangle /\left\langle {d,Gd} \right\rangle ;\quad G = \mu A^T A + \beta D^T D  \label{Eq13}
\end{align}
The two Lagrangian multipliers are then updated by \cite{li2013}: 
\begin{align} 
\upsilon ^{t + 1} &= \upsilon^t - \beta \left( Du^{t + 1} - w^{t + 1} \right)
\label{Eq14} \\ 
\lambda ^{t + 1} &= \lambda^t - \mu \left(Au^{t + 1} - b \right)  \label{Eq15}
\end{align}
Since TV basically assumes piecewise smoothness, it cannot avoid losing detailed information \cite{dong2013compressive}; this is a valid assumption for natural images in smooth regions but is less applicable in non-stationary regions near edges. Consequently, so-called staircase artifacts occur in the recovered image, as shown in Figure~\ref{fig2c}. Moreover, it is worth emphasizing that, even though the signal acquisition is assumed to be noise-free (i.e.,  $Au=b$ ), image signals cannot be perfectly recovered because they cannot be described exactly by $K$-sparse approximation, as shown in \cite{ji2008}. Therefore, if a compressible signal of length $N$ in a selected transform domain is $K$-term approximated by its $K$-largest entries in magnitude $(K < N)$, then the remaining $(N-K)$  elements can be considered as recovery error or noise. By the central limit theorem, it is reasonable to assume that the noise is Gaussian if the sensing matrix is random \cite{ji2008}; in this scenario, the application of a filter will help smoothing the recovered images \cite{gan2007, mun2009, van2013, van2014}. Moreover, for image denoising, the idea of applying a denoising technique to a selected derived image (for example, the normal vectors of the level curves of the noisy image \cite{lysaker2004noise}, the curvature image \cite{bertalmio2014denoising}, or the combined spatial-transformed domain image \cite{knaus2013dual}) might be more effective than directly smoothing the corresponding original noisy image. Motivated by this idea, we suggest that smoothing the Lagrangian multiplier $\upsilon$  can effectively enhance the CS recovered image quality.

\begin{figure}
	\centerline{%
		\setlength\tabcolsep{1.5pt} 
		\begin{tabular}{c c c} 
			\subfloat[]{\label{fig3a} \includegraphics[width= 4cm, height = 4cm]{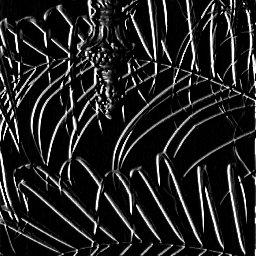}} &
			\subfloat[]{\label{fig3b} \includegraphics[width= 4cm, height = 4cm]{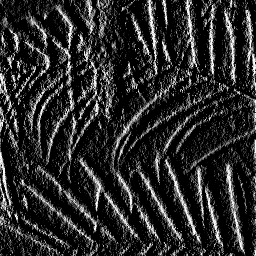}} & 
			\subfloat[]{\label{fig3c} \includegraphics[width= 4cm, height = 4cm]{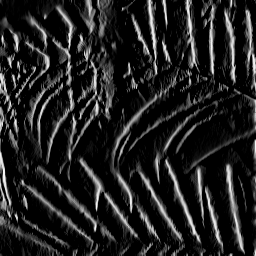}}
		\end{tabular}}
		\caption{Illustration of the Lagrangian multiplier   before and after denoising (block size $16 \times 16$): $(a)$ Subrate $= 1.0$ (i.e., no noise), $(b)$ Subrate $= 0.2$ (before  NLM filter), $(c)$ Subrate $= 0.2$ (after NLM filter)} \label{fig3}
\end{figure}

Lagrangian multipliers are used to find the optimum solution for a multivariate function of CS recovery. The Lagrangian multipliers that represent the gradient image and the measurement vector ($\upsilon$ and $\lambda$, respectively) should have their own roles in solving the ill-posed CS problems. Specially,  $\upsilon$  is updated by the gradient image $Du$ which naturally contains a rich image structure. Hence, in CS recovery, we note that  $\upsilon$  in \eqref{Eq14} can be seen as a noisy version of the gradient image. Indeed, the noise can actually be seen if Figure~\ref{fig3a} and~\ref{fig3b} are compared. With full-Nyquist sampling (i.e., a subrate of $1.0$), there is no noise in $\upsilon$ ; however, there exists a large amount of noise if the subrate is lowered to $0.2$. Also note that, according to the splitting technique, $\upsilon$ plays a role in estimating the solution $u$. Therefore, a more exact $\upsilon$  will provide more accuracy to the $w$ sub-problem and ultimately produce a superior recovered image. Consequently, this suggests the importance of improving the quality of $\upsilon$  in order to obtain better quality with the augmented Lagrangian TV recovery. A proper process should be designed to mitigate noise and artifacts in $\upsilon$. Rather than utilizing Wiener or Gaussian filters, which might easily over-smooth the recovered image \cite{van2013}, we employ the nonlocal means (NLM) filter \cite{buades2005}, which is well-known for its denoising ability while also preserving textures by employing an adaptive weighting scheme with a smoothing parameter that depends on the amount of noise in the signals. The new method to update the Lagrangian multiplier $\upsilon$  is designed as 
\begin{equation}\label{Eq16}
\begin{split} 
\mbox{Step $1$: }& a = \upsilon^t - \beta \left( Du^{t + 1} - w^{t + 1} \right)\\
\mbox{Step $2$: }& \upsilon^{t + 1} = NLM(a)
\end{split}
\end{equation}
Here $NLM (.)$ denotes the NLM filtering operator \cite{buades2005}. We first update $\upsilon$  by the traditional method \cite{li2013} in order to estimate a temporal version (denoted as  $a$ in Step $1$). Next, the NLM filter is applied to reduce the noise in Step $2$. Figure~\ref{fig3c} visualizes the efficiency of the proposed method. After NLM denoising, the Lagrangian multiplier is much cleaner and shows image structures better. Moreover, Figure~\ref{fig2d} shows the recovered image ``Leaves" at a subrate of $0.2$, indicating a reduction of the high-frequency artifacts. The proposed combination of MBTV and the nonlocal Lagrangian multiplier (NLLM) is referred to as MBTV-NLLM and is summarized in Algorithm~\ref{Algorithm1}. 
\begin{algorithm}
		\caption{MBTV-NLLM recovery} \label{Algorithm1}
		\textbf{Input}: Sensing matrix $A$, measurement vector $b$, Lagrangian multipliers, penalty parameters, and $u_0 =A^T b$\\
		\textbf{While} Outer stopping criterion unsatisfied \textbf{do}
		\begin{itemize}
			\item[] \textbf{While} Inner stopping criterion unsatisfied \textbf{do}
			\begin{itemize}
				\item[] Solve the $w$ sub-problem by \eqref{Eq10}
				\item[] Solve the $u$ sub-problem by computing the gradient descent \eqref{Eq11} with  estimation of  the gradient direction via \eqref{Eq12}  and the optimal  step size via \eqref{Eq13} 
			\end{itemize}                         
			\item[]  \textbf{End of while}
			\item[]  Update the multiplier $\upsilon$ using NLM filter by \eqref{Eq16} 
			\item[]  Update the multiplier $\lambda$ by \eqref{Eq15}
		\end{itemize}
		\textbf{End of while}\\
		\textbf{Output}: Final CS recovered image of MBTV-NLLM
\end{algorithm}

At this point, we stress that utilizing the NLM filter for the Lagrangian multiplier yields better recovered images in terms of both subjective and objective qualities. In addition, NLLM has lower computational complexity than nonlocal regularization methods that directly use an NLM filter for noisy images, as in \cite{zhang2010, zhang2013improved}, since the Lagrangian multipliers are only updated if the recovered images are significantly changed. One can refer to Trinh et al. \cite{van2014} for both a theoretical analysis and a numerical comparison between NLLM and nonlocal regularizations \cite{zhang2010, zhang2013improved}. In this paper, we focus on a valid explanation for the gain of the NLLM method, as discussed below.

Mathematically, the error bound of our method (MBTV-NLLM) is smaller than that of traditional TV \cite{li2013} due to the following local convergence statement \cite{bertekas1982constrained}. Assume that  $u^t$ and $\upsilon^t$ are the solutions (see \eqref{Eq11} and \eqref{Eq16}) of the proposed method at the $t$th iteration, where $\tilde{u}^t$  and $\tilde{\upsilon}^t$ are the solutions of TV \cite{li2013}. With a positive scalar $\gamma$, according to proposition $2.4$ \cite{bertekas1982constrained}, the reconstructed errors of MBTV-NLLM and TV \cite{li2013} are 
\begin{align}
\|u^t - \hat{u} \|_1 &\leq \frac{\gamma}{\beta} \|\upsilon^t- \hat{\upsilon} \|_1 \label{Eq17} \\ 
\|\tilde{u}^t - \hat{u} \|_1 &\leq \frac{\gamma}{\beta} \|\tilde{\upsilon}^t- \hat{\upsilon} \|_1  \label{Eq18}
\end{align}
Here, $\hat{\upsilon}$ is the Lagrangian multiplier corresponding to the solution $\hat{u}$.   Hence, we can set up error bounds: 
\begin{align}
\delta _1 &= \frac{\gamma}{\beta}  \| \upsilon^t - \hat{\upsilon} \|_1  \label{Eq19}\\
\delta _2 &= \frac{\gamma}{\beta}  \| \tilde{\upsilon}^t - \hat{\upsilon} \|_1  \label{Eq20}
\end{align}
Note that the solution $\hat{\upsilon}$  prefers to be a clean version (see Figure~\ref{fig3a}). NLLM tries to reduce the noise in $\upsilon^t$  by acquiring the NLM filter (i.e., see \eqref{Eq16}). This implies that $\upsilon^t$  is closer to $\hat{\upsilon}$ than $\tilde{\upsilon}^t $ is (i.e., compare Figure~\ref{fig3b} with Figure~\ref{fig3c}), which can be represented as $\|\upsilon^t- \hat{\upsilon} \|_1 \leq \|\tilde{\upsilon}^t- \hat{\upsilon} \|_1 $. Thus, we obtain
\begin{equation}\label{Eq21}
\delta _1 \leq \delta_2
\end{equation}
The above error coincides with recent reports \cite{needell2013stable}, confirming that the spatial error is bounded by the gradient error. Although the NLM filter provides nonlocal benefits, it still has difficulties in preserving many details in images with low contrast. This drawback is caused by the fact that two very similar pixels on opposite sides produce inaccurate weights for the NLM filter. As a result, some artifacts near edges cannot be sufficiently mitigated without losing detailed information \cite{maleki2012suboptimality}. In this paper, additional processing to enrich the local details through patch-based sparse representation is designed to solve the problem, as proposed in the next sub-section.

\subsection{Refinement for recovered images with patch-based sparse representation }
Classical predetermined transforms, such as DCT or wavelet transforms, cannot always attain a sparse representation of complex details. For example, sharp transitions and singularities in natural images are not expressed well by DCT. In the same way,  $2$D  wavelets might perform poorly for textured or smooth regions \cite{dabov2007}. Recently, for image restoration applications such as denoising, inpainting, or deblurring, patch-based sparse representation has been actively investigated to deal with the complex variations of natural images. Suppose that   $u_i,i = 1, 2, ..., Z,$ (where $Z$ is the total number of patches) denotes an $s \times 1$  column vector representing the $i$th $\sqrt{s} \times \sqrt{s}$ image patch extracted by a patch-extracting operator $R_i$ through $u_i = R_i u$  from an image of size $\sqrt{N} \times \sqrt{N}$  represented by an $N \time 1$ column vector $u$. In this scenario, the image $u$ is synthesized as
\begin{equation}\label{Eq22}
u = \left( \sum_{i = 1}^Z R_i^T R_i  \right)^{- 1}\sum_{i = 1}^Z R_i^T u_i
\end{equation} 
where $(.)^T$  is the regular transpose. Moreover, assume $u_i$  to be sparse over a dictionary $\Phi_i$ with its coefficient vector $\alpha_i$ (that is, $u_i = \Phi_i \alpha_i$). $\Phi$  and $\alpha$  denote the concatenation of dictionaries $\{\Phi_i \}$  and $\{\alpha_i\}$, respectively. Then, $u$ in \eqref{Eq22} is further expressed in a patch-based sparse representation as 
\begin{equation}\label{Eq23}
u = \Phi \circ \alpha = \left( \sum_{i = 1}^Z R_i^T R_i  \right)^{- 1}\sum_{i = 1}^Z R_i^T \Phi_i \alpha_i
\end{equation}  
The operator $\circ$  makes the patch-based sparse representation more compact \cite{elad2006}. Briefly, utilizing a patch-based sparsifying transform to de-correlate the signal and noise in the transform domain is presented by the five following basic steps \cite{dabov2007, dabov2009bm3d, chatterjee2012}:
\begin{enumerate}
	\item Group similar patches: use a nonlocal search to find patches that are similar to the reference patch and stack them in a group.
	\item Forward transform: apply a sparsifying transform (i.e., global or local sparsifying transforms) to each group for transformed coefficients.
	\item Thresholding process: decollate signal and noise by keeping only the significant coefficients. The remaining coefficients are considered to be noise and are discarded. For the CS viewpoint, this step can be referred to as $K$-sparse approximation \cite{gan2007}.
	\item Inverse transform: obtain the estimates for all grouped patches.
	\item Weighting process: return the pixels of the patches to the original locations. The overlapping patches are appropriately weighted according to the number of times each pixel repeats.
\end{enumerate}
For the $K$-sparse approximation (Steps $2$, $3$, and $4$), choosing a proper sparsifying basis will determine the recovered image quality. The authors in \cite{zhang2012image} applied a global transform \cite{dabov2007} that combined $2$D wavelet transform and $1$D DCT transform for all grouped blocks. A predetermined global transform is advantageous in terms of its simplicity; however, it cannot reflect the various sparsity levels of all groups. Therefore, a local transform \cite{dabov2009bm3d, chatterjee2012} was calculated for each individual group to adaptively support the various local sparsity levels. If the local sparsifying transform is poorly designed due to data heavily contaminated by noise, the recovered images will have serious visual artifacts.

Another important quality issue is determining how to collect the proper groups. Related to the nonlocal search discussed in \cite{buades2005, sutour2014adaptive}, patch-based sparse representation still faces some explicit challenges with two outright problems \cite{sutour2014adaptive}: $(i)$ for singular structures, it might fail to find similar patches, thereby producing poor results; and $(ii)$ due to noise, it may detect incorrect patches (i.e., selecting some patches that do not actually belong to the same underlying structure). This can eventually cause over-smoothing. Below, we show that the similarity between a recovered image and its original heavily depends on the variance of error.

\begin{figure}
	\centerline{%
	    \includegraphics[trim=0.0cm 0cm 0cm 0.0cm, clip=true, width=3in]{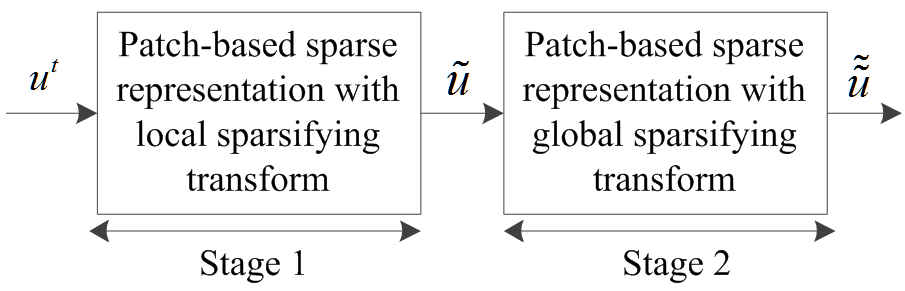}
		}
		\caption{Patch-based sparse representation in two stages} \label{fig4}
\end{figure}

Let us assume that the elements of the error vector $( u - u^{\ast} ) = [ e_1,e_2,...,e_N ]^T$ are independent and come from a normal distribution with zero mean and variance  $\sigma^2.$   Here, $u^{\ast} \in \mathbb{R}^N$  represents a restored version of an original image  $u \in \mathbb{R}^N$  after performing patch-based sparse representation. Since $|e_i|,i = 1,...,N,$  are also independent and come from the half-normal distribution with mean $\sigma \sqrt{2/\pi}$  and variance $\sigma ^2 ( 1 - 2/\pi  )$, a new random variable  $X = ( | e_1 | + |e_2| + \ldots + |e_N | )/N$  has mean and variance of 
\begin{align}
E[X] &= \sigma \sqrt{\frac{2}{\pi }}  \label{Eq24}\\
var [ X ] &= \left( 1 - \frac{2}{\pi } \right)\frac{\sigma^2}{N} \label{Eq25}
\end{align}
Based on the Chebyshev inequality in probability, for a value $\varepsilon > 0$,
\begin{equation}\label{Eq26}
P\left\{ \left| {X - E[ X ]} \right| \leq \varepsilon  \right\}\, \geq 1 - \frac{var \left[ X \right]}{\varepsilon ^2}
\end{equation}
Substituting \eqref{Eq24} and \eqref{Eq25} into \eqref{Eq26}, the probability that expresses the similarity between $u$  and $u^{\ast}$ is 
\begin{equation}\label{Eq27}
P\left\{ \sigma \sqrt{\frac{2}{\pi }}  - \varepsilon  \leq \frac{\|u - u^* \|_1}{N} \leq \sigma \sqrt {\frac{2}{\pi }}  + \varepsilon  \right\} \geq 1 - \left( 1 - \frac{2}{\pi } \right)\frac{\sigma ^2}{N\varepsilon^2}
\end{equation} 
With a sufficiently large image size (i.e., as $N$ becomes large), the probability of similarity between $u$ and $u^{\ast}$ in \eqref{Eq27} approaches $1$. The implication of this is twofold: 
\begin{itemize}
\item First, a good patch-based sparse representation should produce less estimation error (i.e., $\sigma$ is small). 
\item Second, as the noise becomes smaller, the patch-based sparse representation performs better.   
\end{itemize}
The probability of the estimation error in \eqref{Eq27} is important and suggests the idea of combining the local and global sparsifying transforms. At the $t$th iteration, the recovered image $u^t$  is first updated by the five aforementioned steps with a local sparsifying transform. The output of this stage is referred to as $\tilde{u}$, as shown in Figure~\ref{fig4}. Thanks to the local sparsifying transform, $\tilde{u}$ has less noise and fewer artifacts than the input image $u^t$. Additionally, this stage generates a more desirable input version for the second stage, in which we determine the sparsity levels of signals via a global sparsifying transform in order to produce the output image $\tilde{\tilde{u}}$. Our suggested design is described below.
 
\begin{figure}
	\centerline{%
		\includegraphics[trim=3.5cm 1.2cm 8.5cm 4cm, clip=true, width=2.5in]{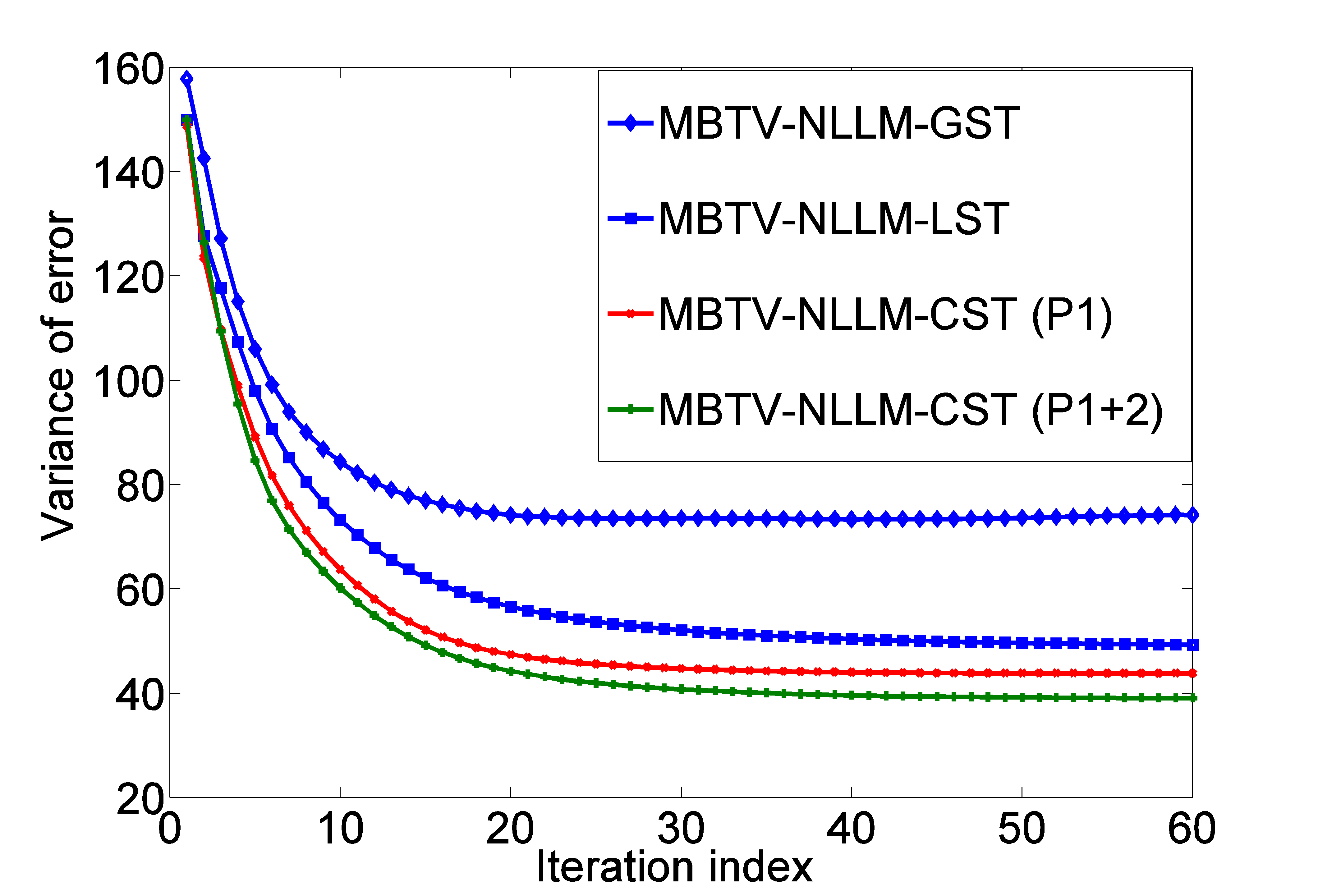}
	}
	\caption{Error reduction over iterations for recovered image Leaves (subrate $0.2$, and block size of $32 \times 32$)} \label{fig5}
\end{figure}

Generally, improving the sparsity level of a signal via a proper transform in CS can be carried out using \cite{xu2014fast, zhang2012image, dong2014compressive, metzler2016denoising, zhang2014image, zhang2014group, van2014block} 
\begin{equation} \label{Eq28}
\underset{\alpha}{\textrm{argmin}} \quad \| \alpha \|_0 \qquad \mbox{s.t. } A\Phi  \circ \alpha  = b
\end{equation}
The unconstrained problem of \eqref{Eq28}, according to the patch-based sparse representation, is formulated to a more tractable optimization problem as 
\begin{equation} \label{Eq29}
\underset{\alpha}{\textrm{argmin}} \left\{ \rho \sum_{i = 1}^Z \|\alpha_i \|_1  + \frac{1}{2} \| b - A\Phi  \circ \alpha  \|_2^2 \right\}
\end{equation}
where $\rho$ is a slack variable ensuring that \eqref{Eq29} is equal to \eqref{Eq28}. We further note that \eqref{Eq29} is a mixed $\ell_1 - \ell_2$  optimization problem that aims at minimizing the cost between the sparsifying coefficients of all of the patches and compressive sensing. For a very simple encoder, the sparsifying transform is moved to the decoder \cite{gan2007}, which means that measurements are directly acquired in the spatial domain (i.e., $Au=b$). According to the modified augmented Lagrangian approach in \cite{afonso2011augmented}, which is a closed form of a sparsifying transform, \eqref{Eq29} is changed to: 
\begin{equation} \label{Eq30}
\underset{u,\alpha}{\textrm{argmin}} \left\{ \rho \sum_{i = 1}^Z \|\alpha_i \|_1  + \frac{1}{2} \| b - Au  \|_2^2  + \frac{\mu_1}{2} \| u - \Phi  \circ \alpha  - \lambda_1 \|_2^2 \right\}
\end{equation}
Here, $\mu_1$  is a positive penalty parameter. The scaled vector $\lambda_1$  is then updated as 
\begin{equation} \label{Eq31}
\lambda _1^{t + 1} = \lambda _1^t - \left( u^{t + 1} - \Phi  \circ \alpha^{t + 1} \right)
\end{equation}
Using the splitting technique \cite{afonso2010fast}, we minimize \eqref{Eq30} by alternatively solving the  $\alpha$  and $u$ sub-problems. More clearly, $\alpha$ is solved by seeking sparsity levels with the five steps of patch-based sparse representation shown in Figure~\ref{fig4}. The $u$ sub-problem is solved by gradient descent $u^{t + 1} = u^t - \eta d$  with an optimized step $\eta$, and the direction $d$ is calculated by the Barzilai$-$Borwein method \cite{li2013, van2013, van2014}: 
\begin{align} 
& d = \mu \left( u^t - \Phi  \circ \alpha ^t - \lambda _1^t \right) - A^T \left( b - Au^t \right) \label{Eq32} \\
&\eta  = \left\langle {d,d} \right\rangle /\left\langle {d,Gd} \right\rangle ;\quad G =  A^T A + \mu I \label{Eq33}
\end{align}
Here, $I$ is an identity matrix. The \underline{c}ombination of local and global \underline{s}parsifying \underline{t}ransforms (CST) and MBTV-NLLM is referred to as MBTV-NLLM-CST. A summary of our proposed recovery method for the BCS framework is depicted in Algorithm~\ref{Algorithm2}. Figure~\ref{fig5} verifies the effectiveness of the suggested design. More clearly, MBTV-NLLM-CST (P$1$) visualizes the reduction in error variance by the first stage in Figure~\ref{fig4} (i.e., using a local sparsifying transform), while MBTV-NLLM-CST (P$1+2$) points out the noise reduction after the two stages in Figure~\ref{fig4}. The gap between the two graphs of MBTV-NLLM-CST (P$1$) and MBTV-NLLM-CST (P$1+2$) indicates the gain from the additional global sparsifying transform. To better understand the nature of this gain, Figure~\ref{fig5} also shows two extra graphs corresponding to MBTV-NLLM with a global sparsifying transform (MBTV-NLLM-GST) and the one with a local sparsifying transform (MBTV-NLLM-LST) only. We recall that the PSNR value is inversely proportional to the variation of error; thus, from Figure~\ref{fig5}, we can confirm that our proposed algorithms always converge to a feasible solution. This is valuable because proving the convergence of a recovery algorithm that deploys a sparsifying transform is not trivial \cite{zhang2014image, zhang2014group}. More interestingly, the results also reveal that the proposed method is better than previous ones \cite{zhang2014image, zhang2014group}, which use either a global sparsifying transform or a local sparsifying transform.
\fontsize{9}{9}{\begin{algorithm} 
	\caption{Proposed recoveries with patch-based sparse representation}  \label{Algorithm2}
	\textbf{Input}: Sensing matrix $A$, measurement vector $b$, Lagrangian multipliers, penalty parameters, and $u_0 =A^T b$\\
	\textcolor{blue}{\textit{\% Initial recovered image by using MBTV-NLLM}}\\
	Call Algorithm~\ref{Algorithm1}\\
	\textcolor{blue}{\textit{\% Sparsity obtained by using global sparsifying transform, local sparsifying transform, or both of them}}\\
	\textbf{While} stopping criterion unsatisfied \textbf{do}
	\begin{itemize}
		\item[] \textbf{If} use global or local sparsifying transform \textbf{do} \textcolor{blue}{\textit{\% MBTV-NLLM-GST or MBTV-NLLM-LST  }}
		\begin{itemize}
			\item[] Solve the $\alpha$ sub-problem by patch-based sparse representation with global or local sparisfying transform
	    \end{itemize}
	    \item[] \textbf{Else} use both global \& local sparsifying transforms \textbf{do} \textcolor{blue}{\textit{\% MBTV-NLLM-CST}} 
	    \begin{itemize}
	    	\item[] Solve the $\alpha$ sub-problem by patch-based sparse representation with local sparsifying transform
	    	\item[] Solve the $\alpha$ sub-problem again by patch-based sparse representation with global sparsifying transform
	    \end{itemize}
	    \item[] \textbf{End if}
	    \item[] Solve the $u$ sub-problem by gradient descent method \eqref{Eq11} with estimation of the gradient direction via \eqref{Eq32} and the optimal step size \eqref{Eq33}.
	    \item[] Update vector $\lambda_1$ by \eqref{Eq31}
	\end{itemize}
	\textbf{End of While}\\
	\textbf{Output}: Final CS recovered image
\end{algorithm}}
\section{Block distributed compressive video sensing} \label{SecIV}
\begin{figure}
	\centerline{%
		\includegraphics[trim=3.5cm 11cm 3cm 10cm, clip=true, width=3.5in]{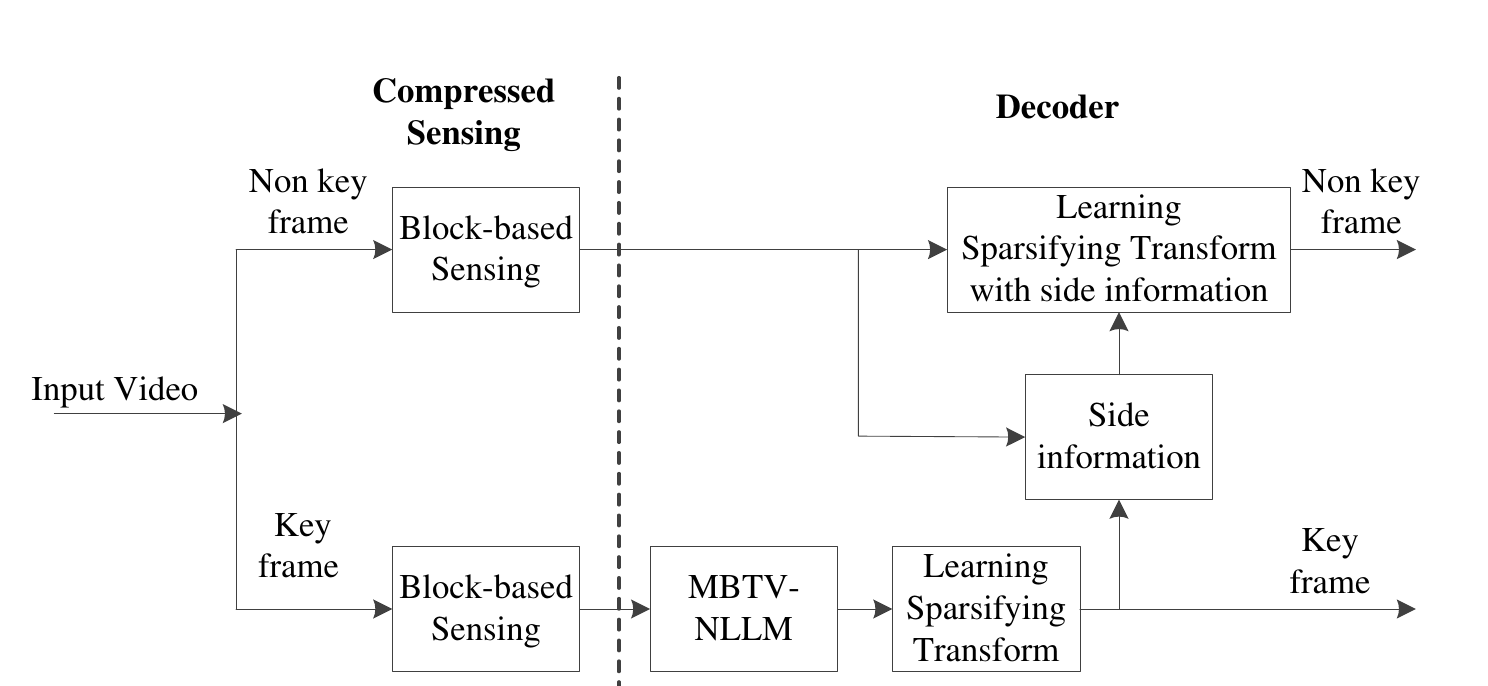}
	}
	\caption{Proposed block DCVS scheme} \label{fig6}
\end{figure}

In this section, we extend the proposed recovery method to block DCVS, as shown in Figure~\ref{fig6}. The main advantage of this design over other existing ones, such as the design proposed in \cite{do2009distributed}, is that it does not require full Nyquist sampling. 
\subsection{Key frame recovery}
A key frame is recovered using the proposed recovery scheme in Algorithm~\ref{Algorithm2}, which was developed for still images. That is, an initial estimate is generated by MBTV-NLLM, and then a sparsifying transform is applied to enrich the local details of the reconstructed key frame.
\subsection{Side information generation}
In distributed video decoding, side information (SI) plays an important role because inaccurate SI strongly degrades the recovery quality of non-key frames. The frames that can be used for reconstructing a non-key frame (denoted by $u_{NK}$) as the side information are 
\begin{equation} \label{Eq34}
\textrm{SI-Frames} = \left\{ u \in G_{\xi} \,|\, \left\| u - u_{NK} \right\|_2 \leq \tau _2 \right\}
\end{equation}
Here, $G_{\xi}$  denotes the $\xi$th  GOP in a video sequence, and the subscript $NK$ indicates a non-key frame. However, the definition in \eqref{Eq34} is impractical when finding proper SI frames for DCVS simply because the only information available at the decoder is the measurement data of non-key frames.  According to the Johnson$-$Lindenstrauss lemma \cite{baraniuk2008simple}, the selection in \eqref{Eq34} can be equivalently written as 
\begin{equation} \label{Eq35}
\textrm{SI-Frames} = \left\{ u \in G_{\xi} \,|\, \| b_{NK} - A_{NK}u \|_2 \leq \tau _2 \right\}
\end{equation}
In a GOP, \eqref{Eq35} allows all of the non-key frames similar to the current non-key frame in the measurement domain to be gathered. The selected SI frames are not much different from each other; thus, an initial non-key frame is computed as their average. Otherwise, it is considered to be a frame with the minimum value of $\| b_{NK} - A_{NK}u \|_2$. Our goal is to find the best initial non-key frame.
\subsection{Recovery of non-key frames}
The non-key frame is refined by its measurement vector, sparsifying transform, and the SI frame denoted by $u_{SI}$: 
 \begin{equation} \label{Eq36}
 \underset{u_{NK},\alpha}{\textrm{argmin}} \left\{  \rho \sum\limits_{i = 1}^Z {{{\left\| {{\alpha _i}} \right\|}_1}}  + \frac{1}{2}\left\| {{b_{NK}} - {A_{NK}}\Phi  \circ \alpha } \right\|_2^2 + \frac{1}{2}\left\| {{u_{NK}} - {u_{SI}}} \right\|_2^2 \right\}
 \end{equation}
Similar to \eqref{Eq30}, \eqref{Eq36} can also be minimized using the augmented Lagrangian approach \cite{maleki2012suboptimality}. This is converted to 
  \begin{equation} \label{Eq37}
  \underset{u_{NK},\alpha}{\textrm{argmin}} \left\{ \rho \sum\limits_{i = 1}^Z {{{\left\| {{\alpha _i}} \right\|}_1}}  + \frac{{{\mu _2}}}{2}\left\| {{u_{NK}} - \Phi  \circ \alpha  - {\lambda _2}} \right\|_2^2 + \frac{1}{2}\left\| {{b_{NK}} - {A_{NK}}{u_{NK}}} \right\|_2^2 + \frac{{{\mu _3}}}{2}\left\| {{u_{NK}} - {u_{SI}} - {\lambda _3}} \right\|_2^2 \right\}
  \end{equation}
Here, $\mu_2$ and $\mu_3$  are positive penalty parameters. At each iteration, the side information is updated, and then the two sub-problems $\alpha$  and $u_{NK}$  are solved. Additionally, the two vectors that represent the sparsifying transform and side information regularization (i.e., $\lambda_2$  and $\lambda_3$, respectively) are updated as 
  \begin{equation} \label{Eq38}
  \left[ {\begin{array}{*{20}{c}}{\lambda _2^{t + 1}}\\{\lambda _3^{t + 1}}\end{array}} \right] = \left[ {\begin{array}{*{20}{c}}{\lambda _2^t}\\{\lambda _3^t}\end{array}} \right] - \left( {\left[ {\begin{array}{*{20}{c}}I\\I\end{array}} \right]{u^{t + 1}} - u_{SI}^{t + 1}} \right)
  \end{equation}
In detail, the side information is first initialized by \eqref{Eq35} and then updated at each iteration by a multi-hypothesis (MH) prediction using Tikhonov regularization \cite{tramel2011video}. Additionally, the $\alpha$  sub-problem is solved by the patch-based sparse representation with a combination of local and global sparsifying transforms, as explained previously. Finally, the $u_{NK}$ sub-problem is solved by gradient descent $u_{NK}^{t + 1} = u_{NK}^t - \eta d$  with the optimal step size and direction estimated by the Barzilai$-$Borwein method \cite{li2013, van2013, van2014}: 
  \begin{align}
   d &= \left\{ {{\mu _1}\left( {u_{NK}^t - \Phi  \circ {\alpha ^t} - \lambda _2^t} \right) + {\mu _2}\left( {u_{NK}^t - {u_{SI}} - \lambda _3^t} \right) - A_{NK}^T\left( {{b_{NK}} - {A_{NK}}{u_{NK}}} \right)} \right\} \label{Eq39}\\
  \eta  &= \left\langle {d,d} \right\rangle /\left\langle {d,Gd} \right\rangle ;G = A_{NK}^T{A_{NK}} + ({\mu _2} + {\mu _3})I  \label{Eq40}
 \end{align}
 \section{Experimental results} \label{SecV}
 \subsection{Test condition}
The recovery performance of the proposed recovery schemes for the BCS framework is evaluated by extensive experiments using both natural images and video. The parameters in the proposed method are experimentally chosen to achieve the best-reconstructed quality. The positive penalty parameters $\theta$  and $\mu$  of MBTV-NLLM are equal to $128$ and $32$, respectively. The NLM filter has a patch size of $7 \times 7$ and a search range of $13 \times 13$. The smoothing parameter is $0.19$. The outer stopping criterion is defined as $\left| \| u^k\|_2 - \| u^{k + 1}\|_2 \right|/\| u^k\|_2 \leq 10^{- 5}$, while the inner stopping criterion is defined as $\left| \| u^k\|_2 - \| u^{k + 1}\|_2 \right|/\| u^k\|_2 \leq 10^{- 4}$. For patch-based sparse representations, the size of the groups is $36 \times 60$, and overlapping is used between patches with an overlapping step size of $2$ pixels. The training window for collecting groups is $30 \times 30$ in size. The penalty parameters $\mu_1$  and $\mu_2$ of the refinement problems are set to  $0.0025$, while $\mu_3$ is set to $0.055$. Additionally, for video recovery, the scale vector $\lambda_1$  is initially set to a zero vector, and the value of  $\tau_2$ is set to $2$. The testing conditions required for the other recovery methods are established based on their suggested recommendations \cite{he2009, he2010, mun2009, chen2011compressed, zhang2014image, zhang2014group}. All of the experiments are performed by MatlabR2011a running on a desktop Intel Corei$3$ RAM$4$G with the Microsoft Windows $7$ operating system. For objective analysis, we use the PSNR (in units of dB). Additionally, the Feature SIMilarity (FSIM) \cite{zhang2011fsim} is used for visual quality evaluation. FSIM is in the range of $[0,1]$, where a value of 1 indicates the best quality.
\begin{table}[ht]
\caption{Performance of various CS recoveries for BCS framework without a patch-based sparse representation. (Unit of PSNR [dB]; $\ast$ denotes the proposed methods)} \label{Table1}
\begin{center}
\scalebox{0.8}{\begin{tabular}{|c|c|c|c|c|c|c|c|c|c|c|c|c|c|}
\hline
\multirow{2}{*}{Image} & \multirow{2}{*}{Subrate} & \multicolumn{2}{c|}{TSDWT\cite{he2009}} & %
\multicolumn{2}{c|}{TSDCT\cite{he2010}} & %
\multicolumn{2}{c|}{SPLCT\cite{mun2009}} & %
\multicolumn{2}{c|}{SPLDDWT\cite{mun2009}} & %
\multicolumn{2}{c|}{MH\cite{chen2011compressed}} & %
\multicolumn{2}{c|}{MBTV-NLLM*}\\
\cline{3-14}
& & PSNR &FSIM &PSNR &FSIM  &PSNR &FSIM &PSNR &FSIM &PSNR &FSIM&PSNR &FSIM\\
\hline
\multirow{4}{*}{Lena} & 0.1 & 22.48 & 0.740 &22.18	&0.767	&24.76	&0.841	&25.31	&0.856	&26.11	& \textbf{0.889}	&\textbf{26.62}	&0.862\\
\cline{2-14}
&0.2 &25.18	&0.839	&26.57	&0.879	&27.48	&0.894	&28.11	&0.906	&\textbf{29.71}	& \textbf{0.933}	&29.31	&0.912\\
\cline{2-14}
&0.3 &27.05	&0.882	&28.73	&0.918	&29.53	&0.924	&30.16	&0.933	&31.26	&\textbf{0.948}	&\textbf{31.29}	&0.941\\
\cline{2-14}
&0.4 &28.50	&0.908	&30.35	&0.940	&31.35	&0.945	&31.98	&0.952	&33.76	&0.966	&33.30	&0.958\\
\hline
\multirow{4}{*}{Leaves} & 0.1 &15.98	&0.589	&17.06	&0.620	&18.56	&0.680	&18.66	&0.685	&20.68	&0.761	&\textbf{21.11}	&\textbf{0.825}\\
\cline{2-14}
&0.2 &18.46	&0.694	&20.68	&0.739	&21.31	&0.758	&21.37	&0.761	&24.81	&0.850	&\textbf{26.13}	&\textbf{0.910}\\
\cline{2-14}
&0.3 &20.51	&0.772	&23.42	&0.807	&23.31	&0.805	&23.30	&0.805	&27.65	&0.898	&\textbf{29.07}	&\textbf{0.941}\\
\cline{2-14}
&0.4 &22.29	&0.827	&25.57	&0.858	&25.31	&0.847	&25.26	&0.847	&29.87	&0.925	&\textbf{31.69}	&\textbf{0.959}\\
\hline
\multirow{4}{*}{Monarch} & 0.1 &18.92	&0.664	&19.71	&0.690	&21.57	&0.772	&21.80	&0.785	&23.68	&0.803	&\textbf{24.54}	&\textbf{0.853}\\
\cline{2-14}
&0.2 &21.50	&0.759	&23.33	&0.782	&24.71	&0.832	&25.26	&0.850	&27.23	&0.869	&\textbf{28.72}	&\textbf{0.915}\\
\cline{2-14}
&0.3 &23.76	&0.818	&25.70	&0.837	&27.10	&0.870	&27.76	&0.885	&29.52	&0.908	&\textbf{31.50}	&\textbf{0.944}\\
\cline{2-14}
&0.4 &25.90	&0.868	&27.68	&0.875	&29.05	&0.898	&29.81	&0.912	&31.28	&0.928	&\textbf{33.47}	&\textbf{0.959}\\
\hline
\multirow{4}{*}{Cameraman} & 0.1 &20.36	&0.685	&20.45	&0.683	&22.12	&0.760	&21.64	&0.762	&22.05	&0.775	&\textbf{23.68}	&\textbf{0.812}\\
\cline{2-14}
&0.2 &22.22	&0.763	&23.05	&0.778	&24.83	&0.825	&24.79	&0.838	&25.41	&0.843	&\textbf{26.68}	&\textbf{0.875}\\
\cline{2-14}
&0.3 &24.36	&0.825	&24.71	&0.825	&26.51	&0.863	&27.02	&0.878	&28.32	&0.898	&\textbf{28.44}	&\textbf{0.912}\\
\cline{2-14}
&0.4 &25.97	&0.863	&26.60	&0.869	&28.41	&0.894	&28.99	&0.909	&29.81	&0.918	&\textbf{30.38}	&\textbf{0.937}\\
\hline
\multirow{4}{*}{House} & 0.1 &23.91	&0.719	&24.75	&0.767	&26.69	&0.836	&26.95	&0.846	&30.07	&\textbf{0.895}	&\textbf{30.47}	&0.877\\
\cline{2-14}
&0.2 &27.20	&0.837	&29.23	&0.871	&29.95	&0.894	&30.56	&0.902	&\textbf{33.73}	&\textbf{0.938}	&33.51	&0.920\\
\cline{2-14}
&0.3 &29.64	&0.884	&32.03	&0.912	&32.34	&0.926	&32.83	&0.931	&\textbf{35.62}	&\textbf{0.956}	&35.58	&0.945\\
\cline{2-14}
&0.4 &32.30	&0.920	&34.09	&0.938	&34.18	&0.947	&34.67	&0.949	&\textbf{37.20}	&\textbf{0.967}	&36.98	&0.958\\
\hline
\multirow{4}{*}{Parrot} &0.1 &21.93	&0.755	&22.58	&0.830	&23.26	&0.865	&23.32	&0.876	&24.29	&\textbf{0.887}	&\textbf{25.15}	&0.880\\
\cline{2-14}
&0.2 &24.25	&0.849	&25.58	&0.895	&26.11	&0.909	&26.36	&0.918	&\textbf{27.92}	&\textbf{0.928}	&27.91	&0.922\\
\cline{2-14}
&0.3 &25.54	&0.890	&27.48	&0.926	&28.29	&0.935	&28.63	&0.943	&\textbf{30.79}	&\textbf{0.952}	&30.33	&0.942\\
\cline{2-14}
&0.4 &27.11	&0.918	&28.91	&0.942	&30.20	&0.952	&30.92	&0.958	&32.54	&\textbf{0.964}	&\textbf{32.92}	&0.957\\
\hline
\multirow{4}{*}{Boat} & 0.1 &21.78	&0.679	&19.82	&0.722	&24.22	&0.799	&24.52	&0.802	&26.12	&\textbf{0.852}	&\textbf{26.76}	&0.844\\
\cline{2-14}
&0.2 &24.75	&0.800	&26.47	&0.848	&26.94	&0.864	&27.08	&0.866	&\textbf{30.17}	&\textbf{0.920}	&30.06	&0.913\\
\cline{2-14}
&0.3 &26.57	&0.854	&28.84	&0.900	&29.05	&0.903	&28.97	&0.900	&32.42	&\textbf{0.944}	&\textbf{32.54}	&0.943\\
\cline{2-14}
&0.4 &28.73	&0.901	&31.44	&0.936	&30.88	&0.929	&30.61	&0.925	&34.14	&0.960	&\textbf{34.72}	&\textbf{0.963}\\
\hline
\multirow{4}{*}{Pepper} & 0.1 &20.37	&0.695	&21.02	&0.739	&23.67	&0.826	&24.37	&0.838	&25.78	&\textbf{0.859}	&\textbf{25.87}	&\textbf{0.859}\\
\cline{2-14}
&0.2 &22.50	&0.785	&25.00	&0.836	&27.03	&0.883	&27.63	&0.890	&29.32	&0.910	& \textbf{30.28}	&\textbf{0.921}\\
\cline{2-14}
&0.3 &26.20	&0.870	&28.51	&0.901	&29.17	&0.910	&29.82	&0.918	&31.20	&0.933	&\textbf{32.75}	&\textbf{0.947}\\
\cline{2-14}
&0.4 &29.49	&0.912	&30.98	&0.931	&30.98	&0.932	&31.72	&0.939	&32.91	&0.950	& \textbf{34.62}	&\textbf{0.962}\\
\hline
\multicolumn{2}{|c|}{Average} &24.24	&0.805	&25.70	&0.836	&26.84	&0.866	&27.19	&0.874	&29.23	&0.904	&\textbf{29.89}	&\textbf{0.915}\\
\hline
\multicolumn{2}{|c|}{Gain by  MBTV-NLLM$^{\ast}$} &5.65	&0.110	&4.19	&0.079	&3.05	&0.049	&2.70	&0.041	&0.66	&0.011	&-	&-\\
\hline
\end{tabular}}

\end{center}
\end{table}


\begin{table}[ht]
	\caption{Performance of various CS recoveries for BCS framework with a patch-based sparse representation. (Unit of PSNR [dB]; $\ast$ denotes the proposed methods)} \label{Table2}
	\begin{center}
		\scalebox{0.8}{\begin{tabular}{|c|c|c|c|c|c|c|c|c|c|c|c|c|c|}
				\hline
				\multirow{2}{*}{Image} & \multirow{2}{*}{Subrate} & \multicolumn{2}{c|}{RALS\cite{zhang2014image}} & %
				\multicolumn{2}{c|}{GSR\cite{zhang2014group}} & %
				\multicolumn{2}{c|}{\vtop{\hbox{\strut MBTV-NLLM}\hbox{\strut-GST$^\ast$}}} & %
				\multicolumn{2}{c|}{\vtop{\hbox{\strut MBTV-NLLM}\hbox{\strut-LST$^\ast$}}} & %
				\multicolumn{2}{c|}{\vtop{\hbox{\strut MBTV-NLLM}\hbox{\strut-CST$^\ast$}}} \\
				\cline{3-12}
				& & PSNR &FSIM &PSNR &FSIM  &PSNR &FSIM &PSNR &FSIM &PSNR &FSIM\\
				\hline
				\multirow{4}{*}{Lena} & 0.1 & 27.07	&0.899	&27.57	&0.915	&27.63	&0.914	&\textbf{28.25}	&\textbf{0.915}	&28.02	&0.914\\
				\cline{2-12}
				&0.2 &30.49	&0.943	&30.88	&0.952	&30.98	&0.951	&31.57	&0.952	&\textbf{31.66}	&\textbf{0.954}\\
				\cline{2-12}
				&0.3 &33.17	&0.966	&33.96	&0.971	&33.51	&0.969	&34.13	&0.970	&\textbf{34.51}	&\textbf{0.972}\\
				\cline{2-12}
				&0.4 &35.50	&0.978	&36.46	&0.981	&35.90	&0.979	&36.21	&0.980	&\textbf{36.71}	&\textbf{0.981}\\
				\hline
				\multirow{4}{*}{Leaves} & 0.1 &21.55	&0.801	&23.22	&0.876	&23.69	&0.890	&25.68	&0.914	&\textbf{26.55}	&\textbf{0.926}\\
				\cline{2-12}
				&0.2 &27.13	&0.909	&30.54	&0.956	&29.48	&0.950	&31.23	&0.961	&\textbf{32.21}	&\textbf{0.967}\\
				\cline{2-12}
				&0.3 &31.20	&0.951	&34.40	&0.976	&33.32	&0.972	&34.81	&0.978	&\textbf{36.13}	&\textbf{0.983}\\
				\cline{2-12}
				&0.4 &34.69	&0.974	&37.63	&0.987	&36.19	&0.983	&37.69	&0.987	&\textbf{39.00}	&\textbf{0.990}\\
				\hline
				\multirow{4}{*}{Monarch} & 0.1 &24.42	&0.827	&25.28	&0.867	&25.88	&0.882	&27.73	&0.912	&\textbf{28.08}	&\textbf{0.915}\\
				\cline{2-12}
				&0.2 &28.36	&0.892	&30.77	&0.941	&30.27	&0.931	&31.98	&0.953	&\textbf{32.80}	&\textbf{0.956}\\
				\cline{2-12}
				&0.3 &31.60	&0.933	&34.25	&0.964	&33.34	&0.956	&35.04	&0.971	&\textbf{35.99}	&\textbf{0.974}\\
				\cline{2-12}
				&0.4 &34.39	&0.957	&36.86	&0.976	&35.58	&0.969	&37.35	&0.980	&\textbf{38.32}	&\textbf{0.982}\\
				\hline
				\multirow{4}{*}{Cameraman} & 0.1 &22.93	&0.799	&22.90	&0.815	&24.71	&0.849	&\textbf{25.58}	&\textbf{0.860}	&25.38	&0.853\\
				\cline{2-12}
				&0.2 &26.59	&0.875	&26.76	&0.889	&28.17	&0.912	&28.74	&\textbf{0.915}	&\textbf{28.78}	&\textbf{0.915}\\
				\cline{2-12}
				&0.3 &29.26	&0.922	&28.97	&0.928	&30.60	&0.938	&30.90	&0.940	&\textbf{31.22}	&\textbf{0.944}\\
				\cline{2-12}
				&0.4 &31.19	&0.945	&31.25	&0.953	&32.39	&0.956	&32.42	&0.956	&\textbf{32.85}	&\textbf{0.961}\\
				\hline
				\multirow{4}{*}{House} & 0.1 &32.10	&0.911	&32.94	&0.920	&33.31	&0.928	&33.17	&0.927	&\textbf{33.88}	&\textbf{0.930}\\
				\cline{2-12}
				&0.2 &36.08	&0.955	&\textbf{37.37}	&\textbf{0.964}	&36.43	&0.959	&36.37	&0.958	&37.19	&0.963\\
				\cline{2-12}
				&0.3 &38.37	&0.973	&\textbf{39.41}	&\textbf{0.978}	&38.81	&0.975	&38.41	&0.974	&39.20	&0.977\\
				\cline{2-12}
				&0.4 &40.17	&0.982	&\textbf{40.90}	&0.984	&40.32	&0.982	&40.03	&0.982	&40.88	&\textbf{0.985}\\
				\hline
				\multirow{4}{*}{Parrot} & 0.1 &25.33	&0.908	&25.98	&0.923	&26.28	&0.919	&27.41	&0.921	&\textbf{27.47}	&\textbf{0.924}\\
				\cline{2-12}
				&0.2 &29.34	&0.944	&30.76	&0.952	&29.54	&0.949	&\textbf{31.75}	&\textbf{0.953}	&31.15	&\textbf{0.953}\\
				\cline{2-12}
				&0.3 &32.49	&0.963	&\textbf{34.17}	&\textbf{0.968}	&32.39	&0.963	&34.15	&0.966	&34.15	&0.967\\
				\cline{2-12}
				&0.4 &34.78	&0.974	&36.61	&\textbf{0.978}	&34.99	&0.975	&36.34	&0.976	&\textbf{36.84}	&\textbf{0.978}\\
				\hline
				\multirow{4}{*}{Boat} &0.1 &28.01	&0.890	&28.28	&0.901	&28.69	&0.901	&29.11	&\textbf{0.908}	&\textbf{29.18}	&0.907\\
				\cline{2-12}
				&0.2 &33.01	&0.952	&33.69	&0.958	&33.34	&0.956	&33.31	&0.955	&\textbf{34.02}	&\textbf{0.960}\\
				\cline{2-12}
				&0.3 &36.29	&0.973	&36.67	&0.975	&36.14	&0.973	&36.00	&0.972	&\textbf{36.90}	&\textbf{0.978}\\
				\cline{2-12}
				&0.4 &38.89	&0.984	&\textbf{39.26}	&\textbf{0.985}	&38.48	&0.983	&38.50	&0.983	&38.94	&\textbf{0.985}\\
				\hline
				\multirow{4}{*}{Pepper} &0.1 &27.54	&0.892	&28.18	&0.908	&28.91	&0.913	&28.74	&0.911	&\textbf{29.44}	&\textbf{0.917}\\
				\cline{2-12}
				&0.2 &31.64	&0.941	&32.59	&0.951	&32.73	&0.953	&32.56	&0.952	&\textbf{33.28}	&\textbf{0.955}\\
				\cline{2-12}
				&0.3 &34.39	&0.963	&35.07	&0.966	&35.12	&0.967	&34.81	&0.965	&\textbf{35.68}	&\textbf{0.969}\\
				\cline{2-12}
				&0.4 &36.52	&0.974	&37.00	&0.976	&37.03	&\textbf{0.978}	&36.56	&0.975	&\textbf{37.42}	&\textbf{0.978}\\
				\hline
				\multicolumn{2}{|c|}{Average} &31.39	&0.930	&32.52	&0.945	&32.32	&0.946	&33.02	&0.951	&\textbf{33.56}	&\textbf{0.954}\\
				\hline
				\multicolumn{2}{|c|}{Gain by  MBTV-NLLM-CST$^{\ast}$} &2.17	&0.024	&1.04	&0.009	&1.24	&0.008	&0.54	&0.003	&-	&-\\
				\hline
			\end{tabular}}
		\end{center}
	\end{table}
\subsection{Test results with still images}
\begin{figure}[t]
	\centerline{%
		\setlength\tabcolsep{1.5pt} 
		\begin{tabular}{c c c c} 
			\includegraphics[width= 2.5cm, height =2.5cm]{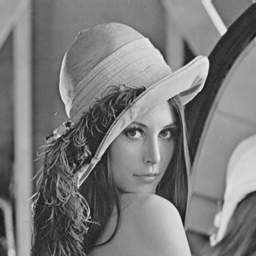} &
			\includegraphics[width= 2.5cm, height = 2.5cm]{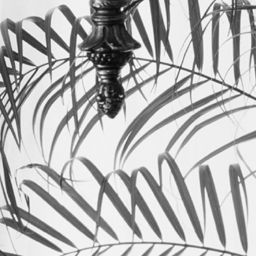} & \includegraphics[width=2.5cm, height = 2.5cm]{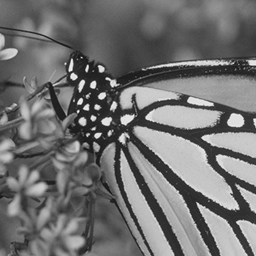} & \includegraphics[width= 2.5cm, height = 2.5cm]{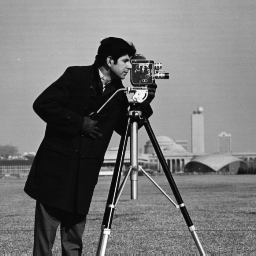} \\
			\includegraphics[width= 2.5cm, height = 2.5cm]{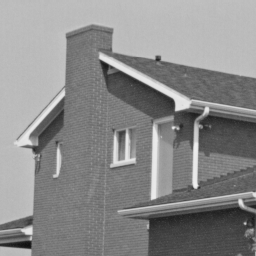} &
			\includegraphics[width=2.5cm, height = 2.5cm]{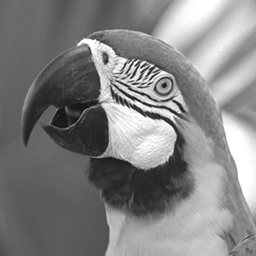} & \includegraphics[width= 2.5cm, height = 2.5cm]{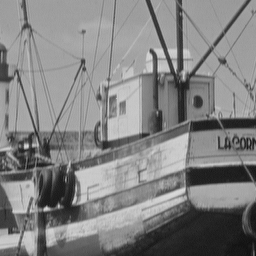} & \includegraphics[width= 2.5cm, height = 2.5cm]{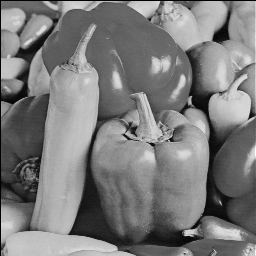}
		\end{tabular}}
		\caption{Original tested images} \label{fig7}
	\end{figure}

Eight well-known $256 \times 256$ natural images are used, that is, Lena, Leaves, Monarch, Cameraman, House, Boat, and Pepper, as shown in Figure~\ref{fig7}. For fair comparisons with previous works \cite{he2009, he2010, mun2009,chen2011compressed, zhang2014image, zhang2014group}, the natural images are divided into non-overlapping blocks ($32 \times 32$ in size). They are compressively sensed by an i.i.d. random Gaussian sensing matrix. Table~\ref{Table1} compares five well-known existing CS recovery methods (i.e., the tree-structured CS with variational Bayesian analysis using DWT (TSDWT) \cite{he2009}, the tree-structured CS with variational Bayesian analysis using DCT (TSDCT) \cite{he2010}, SPLDDWT \cite{mun2009}, the SPL using the contourlet transform (SPLCT) \cite{mun2009}, and the multi-hypothesis CS method (MH) \cite{chen2011compressed}) with the proposed MBTV-NLLM when patch-based sparse representation is not employed. It is worth emphasizing that, for this test case, MH is by far the most state-of-the-art method. However, it turns out that the proposed MBTV-NLLM is competitive with MH and much better than the others. In the best case, MBTV-NLLM surpasses TSDWT, TSDCT, SPLCT, SPLDWT, and MH by $9.40$ dB, $6.94$ dB, $6.38$ dB, $6.43$ dB, and $2.19$ dB, respectively. Thus, it successfully demonstrates the effectiveness of the proposed schemes: MBTV and denoising of the Lagrange multiplier. The last rows of Tables~\ref{Table1} and ~\ref{Table2} show the gains achieved by the proposed MBTV-NLLM and MBTV-NLLM-CST, respectively, with respect to each individual method.

\begin{figure}[t]
	\centerline{%
		\setlength\tabcolsep{1.5pt} 
		\begin{tabular}{c c c} 
			\subfloat[Ground Truth]{\includegraphics[width= 4cm, height = 4cm]{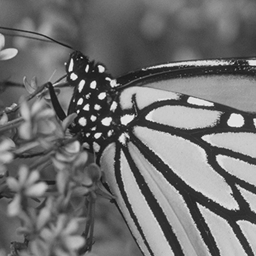}} &
			\subfloat[TSDWT\cite{he2009}]{\includegraphics[width= 4cm, height = 4cm]{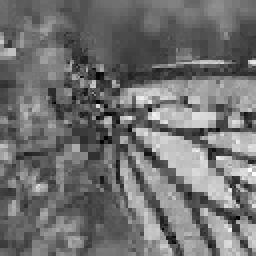}} & 
			\subfloat[TSDCT\cite{he2010}]{\includegraphics[width= 4cm, height = 4cm]{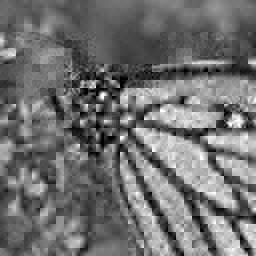}} \vspace{-0.3cm} \\
			\subfloat[SPLCT\cite{mun2009}]{\includegraphics[width= 4cm, height = 4cm]{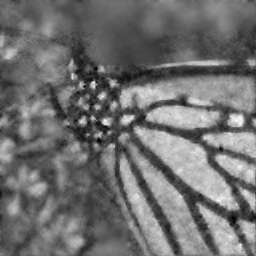}} &
			\subfloat[SPLDDWT\cite{mun2009}]{\includegraphics[width= 4cm, height = 4cm]{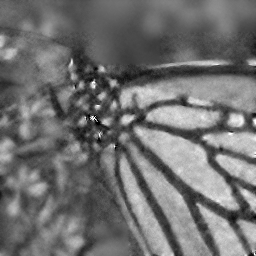}} &
			 \subfloat[MH\cite{chen2011compressed}]{\includegraphics[width= 4cm, height = 4cm]{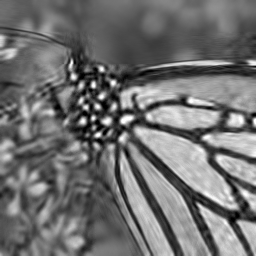}} \vspace{-0.3cm}\\
			\subfloat[MBTV-NLLM$^\ast$]{\includegraphics[width= 4cm, height = 4cm]{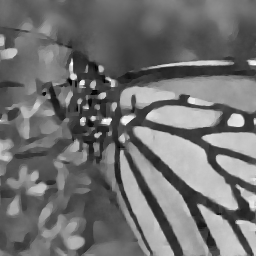} }&
			\subfloat[RALS\cite{zhang2014image}]{\includegraphics[width= 4cm, height = 4cm]{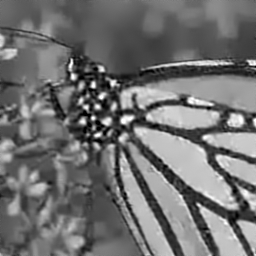}} & 
			\subfloat[GSR\cite{zhang2014group}]{\includegraphics[width= 4cm, height = 4cm]{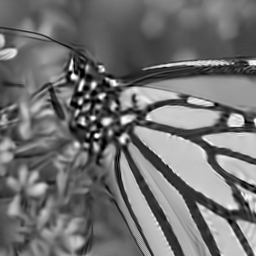}} \vspace{-0.3cm}\\
			\subfloat[MBTV-NLLM-GST*]{\includegraphics[width= 4cm, height = 4cm]{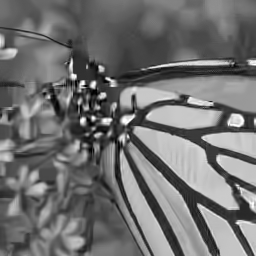}} &
			\subfloat[MBTV-NLLM-LST*]{\includegraphics[width= 4cm, height = 4cm]{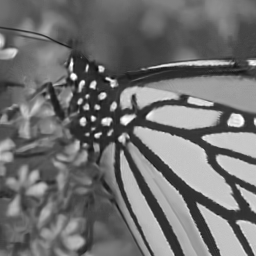}} &
			\subfloat[MBTV-NLLM-CST*]{\includegraphics[width= 4cm, height = 4cm]{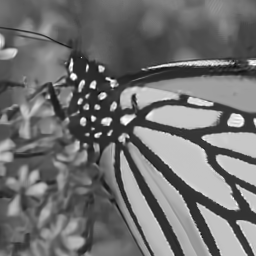}} \vspace{-0.3cm}
		\end{tabular}}
		\caption{Visual quality comparison of various CS recovery methods for BCS framework (subrate $0.1$, block-size $32 \times 32$, *denotes the proposed methods)} \label{fig8}
\end{figure}

Further effectiveness of the proposed patch-based sparse representation is demonstrated in Table~\ref{Table2}. In \cite{zhang2014image}, the authors employed KSVD \cite{elad2006} to design a recovery method that used adaptively-learned sparsifying (RALS), while the group sparse representation (GSR) \cite{zhang2014group} acquired the local sparsifying transform. GSR is certainly better than RALS because of the local sparsifying basis for each group. In our three proposed methods in Table~\ref{Table2}, MBTV-NLLM-GST attains better reconstructed quality than RALS, while MBTV-NLLM-LST and MBTV-NLLM-CST outperform GSR. This is because the better initial image created by MBTV-NLLM has beneficial effects on grouping patches by facilitating a better non-local search and defining more appropriate sparsifying bases for each group (when using a local sparsifying transform). In particular, the PSNR of MBTV-NLLM-CST is as much as $3.33$ dB higher than GSR for the recovered image Leaves.

\begin{figure}
	\centerline{%
		\setlength\tabcolsep{1.5pt} 
		\begin{tabular}{c c c} 
			\subfloat[Ground Truth]{\includegraphics[width= 4cm, height = 4cm]{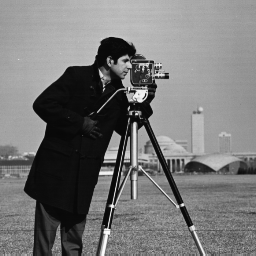}} &
			\subfloat[TSDWT\cite{he2009}]{\includegraphics[width= 4cm, height = 4cm]{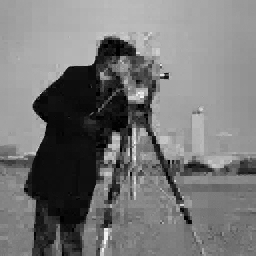}} & 
			\subfloat[TSDCT\cite{he2010}]{\includegraphics[width= 4cm, height = 4cm]{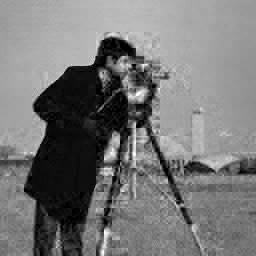}} \vspace{-0.3cm} \\
			\subfloat[SPLCT\cite{mun2009}]{\includegraphics[width= 4cm, height = 4cm]{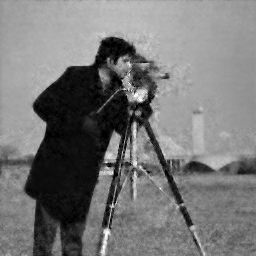}} &
			\subfloat[SPLDDWT\cite{mun2009}]{\includegraphics[width= 4cm, height = 4cm]{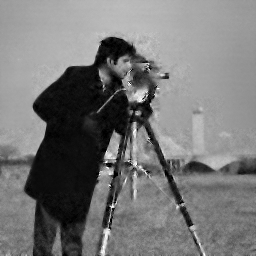}} &
			\subfloat[MH\cite{chen2011compressed}]{\includegraphics[width= 4cm, height = 4cm]{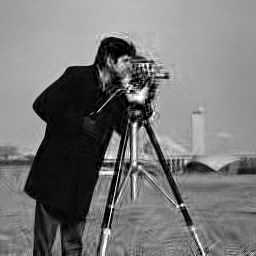}} \vspace{-0.3cm} \\
			\subfloat[MBTV-NLLM$^\ast$]{\includegraphics[width= 4cm, height = 4cm]{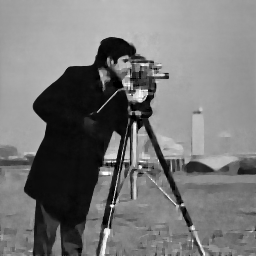}} &
			\subfloat[RALS\cite{zhang2014image}]{\includegraphics[width= 4cm, height = 4cm]{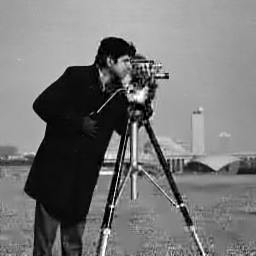}} &
			\subfloat[GSR\cite{zhang2014group}]{\includegraphics[width= 4cm, height = 4cm]{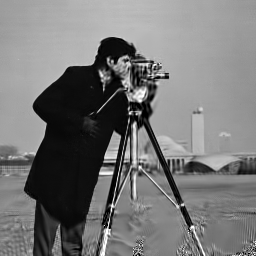}} \vspace{-0.3cm}\\
			\subfloat[MBTV-NLLM-GST*]{\includegraphics[width= 4cm, height = 4cm]{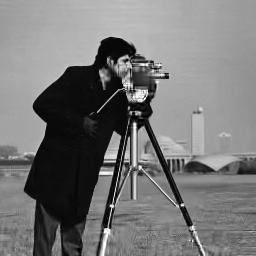}} &
			\subfloat[MBTV-NLLM-LST*]{\includegraphics[width= 4cm, height = 4cm]{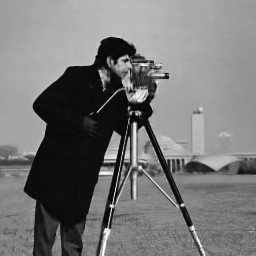}} & 
			\subfloat[MBTV-NLLM-CST*]{\includegraphics[width= 4cm, height = 4cm]{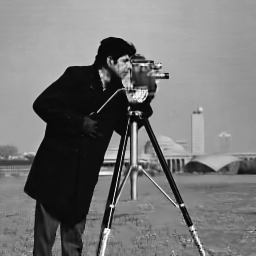}} \vspace{-0.3cm}
		\end{tabular}}
		\caption{Visual quality of various CS recoveries  for BCS framework
			(subrate $0.2$, block-size $32 \times 32$, *denotes the proposed methods)} \label{fig9}
	\end{figure}

Furthermore, for a complex image with as much detail as is found in the image Lena, MBTV-NLLM-CST is not as successful as MBTV-NLLM-GST at a subrate of $0.1$. The recovered image lacks spatial detail at a very low-subrate such that the combination of local and global sparsifying transforms might make it slightly over-smoothed. The visual quality of the proposed schemes and previous work are compared in Figure~\ref{fig8} and Figure~\ref{fig9} using the image Monarch at subrate $0.1$ and Cameraman at subrate $0.2$. This test shows that, while all conventional CS recovery schemes \cite{he2009, he2010, mun2009, chen2011compressed, zhang2014image, zhang2014group} suffer from a large degree of high-frequency artifacts, including the state-of-art method (GSR), the three proposed schemes seem to work much better. However, the recovered image of MBTV-NLLM-GST still has some artifacts at a very low subrate (e.g., see the image Monarch at subrate $0.1$). This indicates that a global sparsifying transform cannot adequately express the sparsifying levels for all groups.

\begin{figure} [t]
 	\centerline{%
 		\setlength\tabcolsep{1.5pt} 
 		\begin{tabular}{c c c} 
 			\subfloat[Lena]{\includegraphics[trim=6cm 1cm 8.5cm 4.8cm, clip=true, width=2.3in]{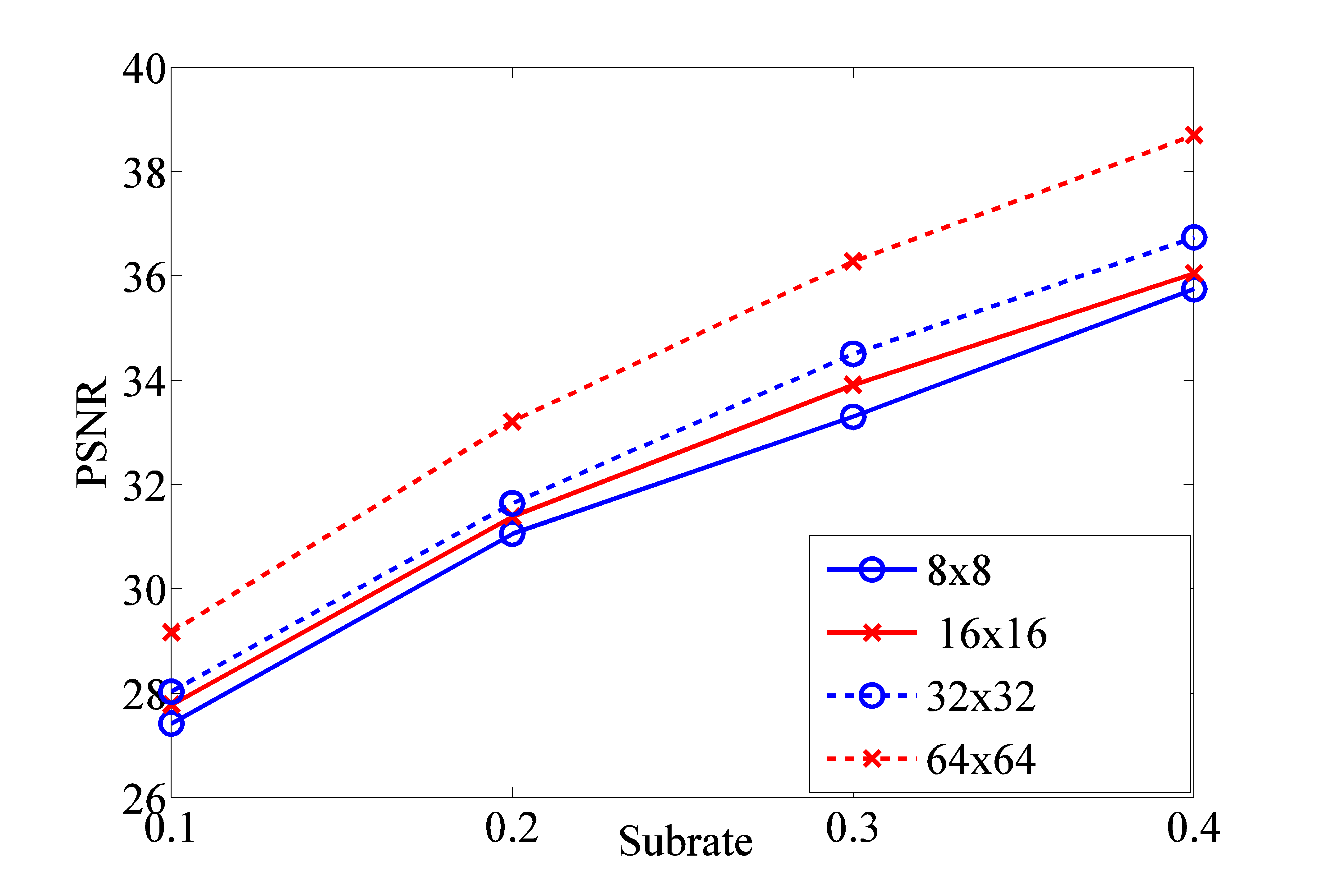}} &
 			\subfloat[Leaves]{\includegraphics[trim=6cm 1cm 8.5cm 4.8cm, clip=true, width=2.3in]{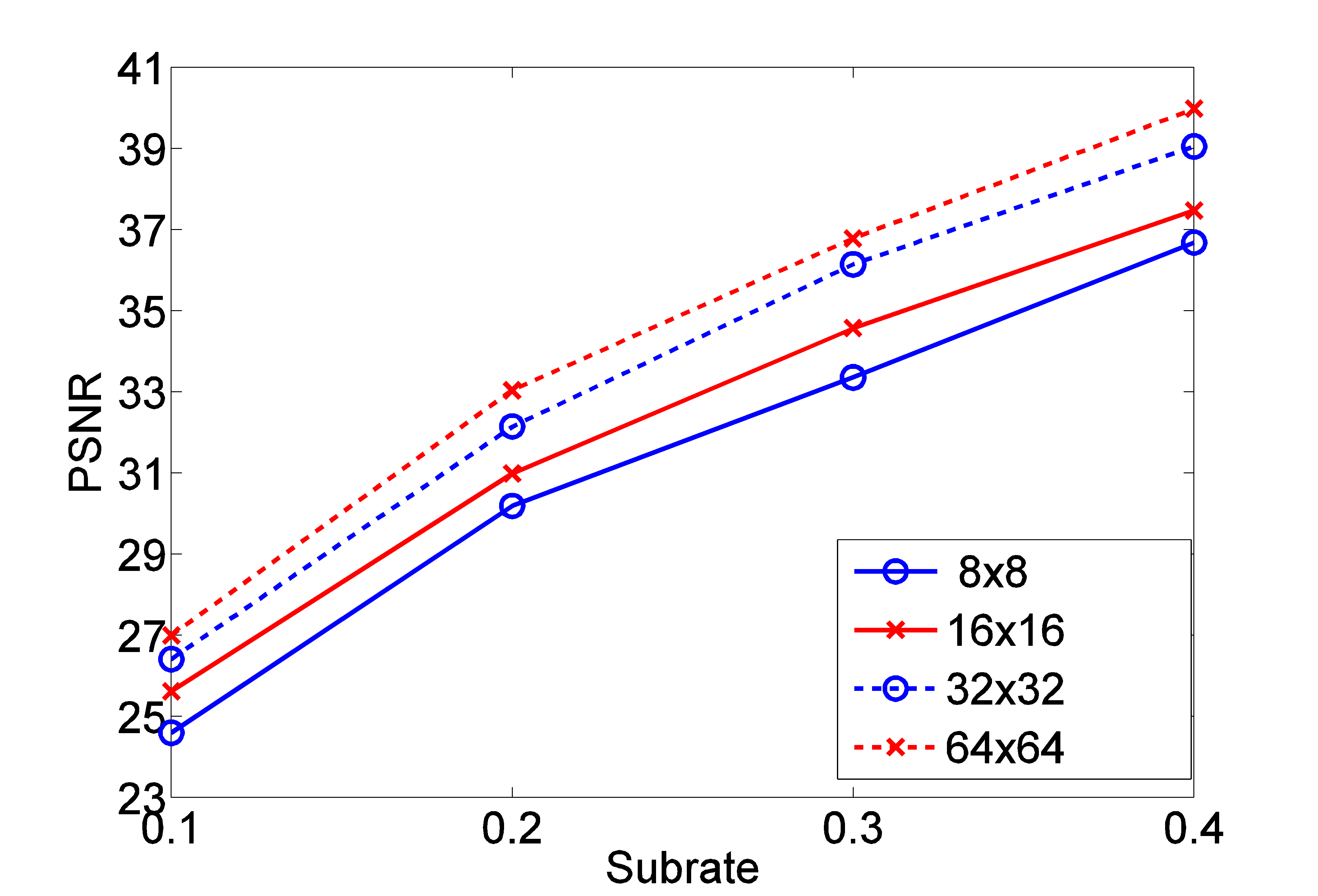}} &
 			 \subfloat[Cameraman]{\includegraphics[trim=6cm 1cm 8.5cm 4.8cm, clip=true, width=2.3in]{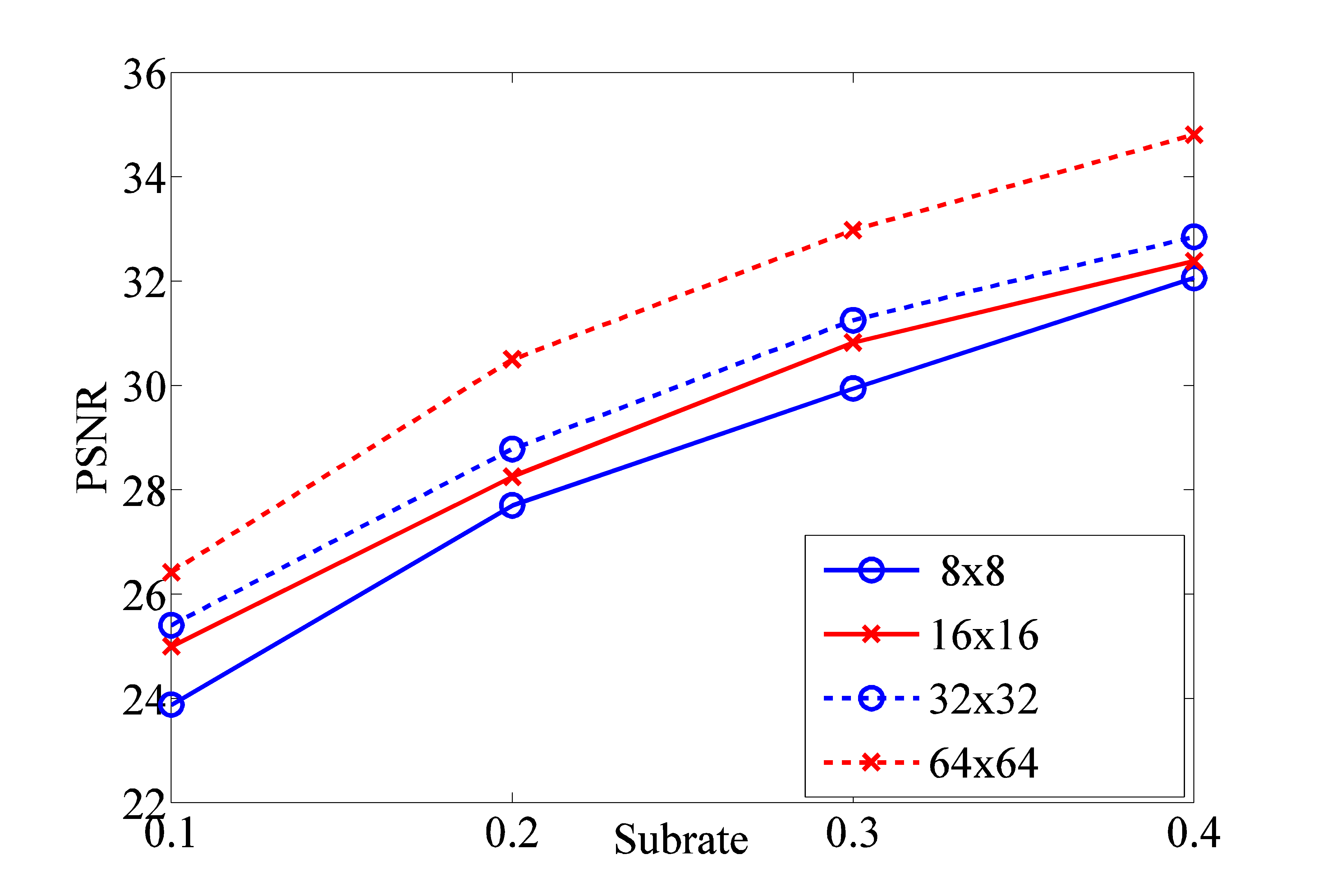}}
 		\end{tabular}}
 		\caption{PSNR versus various block size ($8 \times 8, 16 \times 16, 32 \times 32,$ and $64 \times 64$)} \label{fig10}
\end{figure}

Figure~\ref{fig10} quantifies the effectiveness of MBTV-NLLM-CST according to block sizes utilizing three images (i.e., Lena, Leaves, and Cameraman). Increasing the size of the sensing matrix yields better quality in the recovered images in terms of the PSNR. For example, at subrate $0.1$ with a block size of $8 \times 8$, the recovered image Lena has a PSNR of $24.60$  dB. However, this value can be increased up to $27.00$ dB with a block size of $64 \times 64$. These results coincide with our analysis in Section II based on the RIP property.

\subsection{Test results with video for block DCVS}

\begin{figure*}[t]
 	\centerline{%
 		\setlength\tabcolsep{1.5pt} 
 		\begin{tabular}{c c c} 
 			\subfloat[News]{\includegraphics[trim=4cm 1cm 8.5cm 4.8cm, clip=true, width=2.3in]{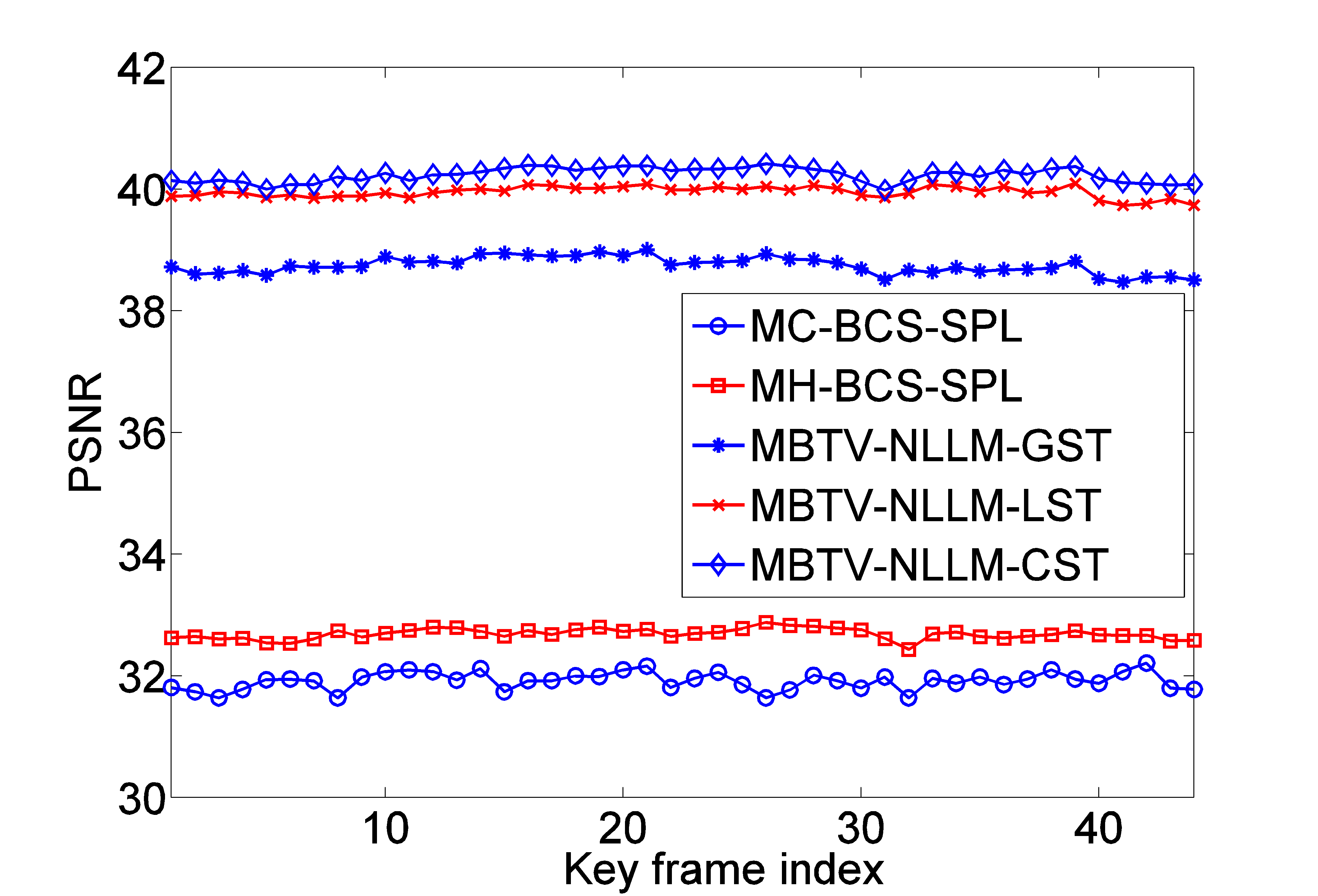}} &
 			\subfloat[Mother-daughter]{\includegraphics[trim=4cm 1cm 8.5cm 4.8cm, clip=true, width=2.3in]{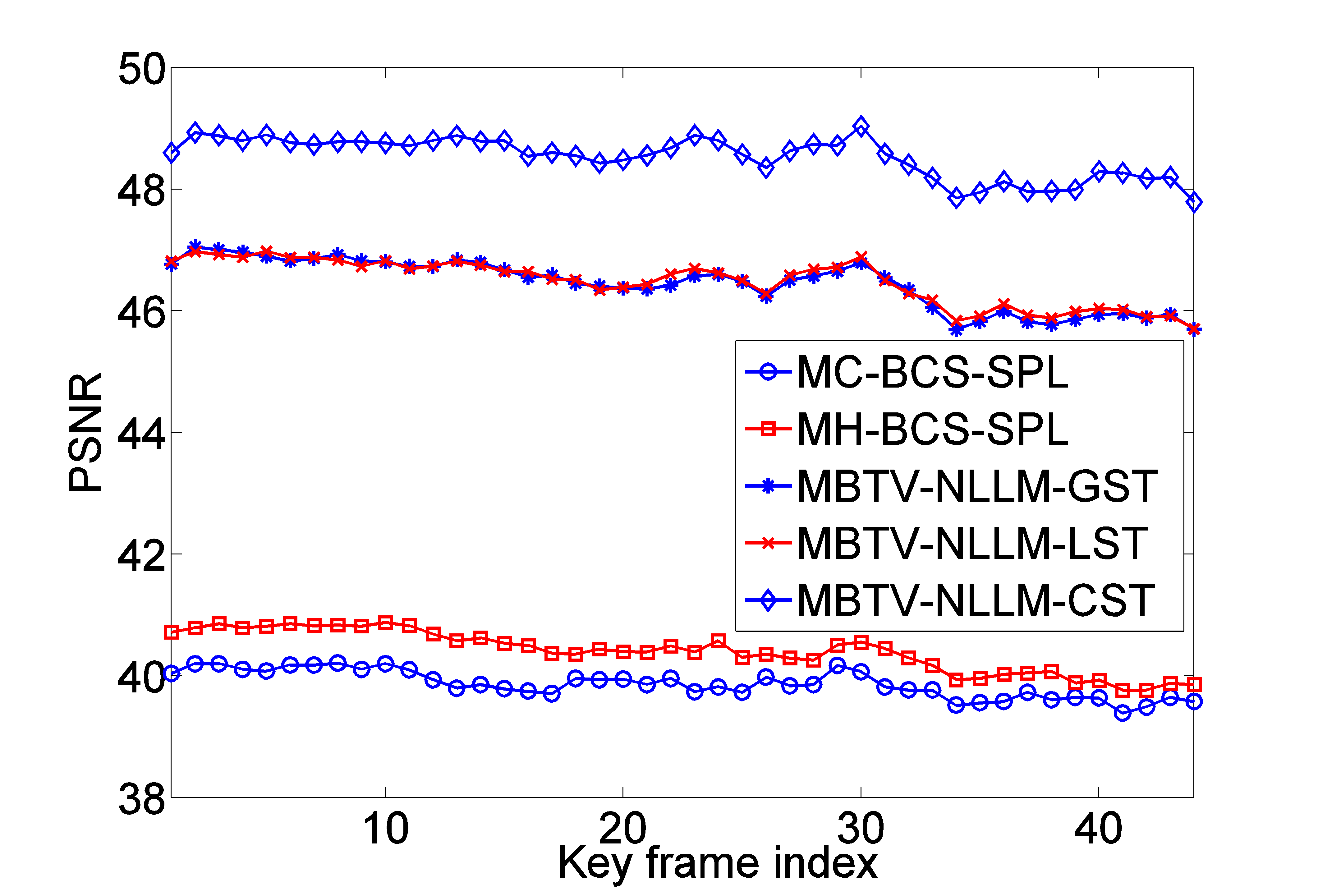}} & \subfloat[Salesman]{\includegraphics[trim=3cm 1cm 8.5cm 4.8cm, clip=true, width=2.3in]{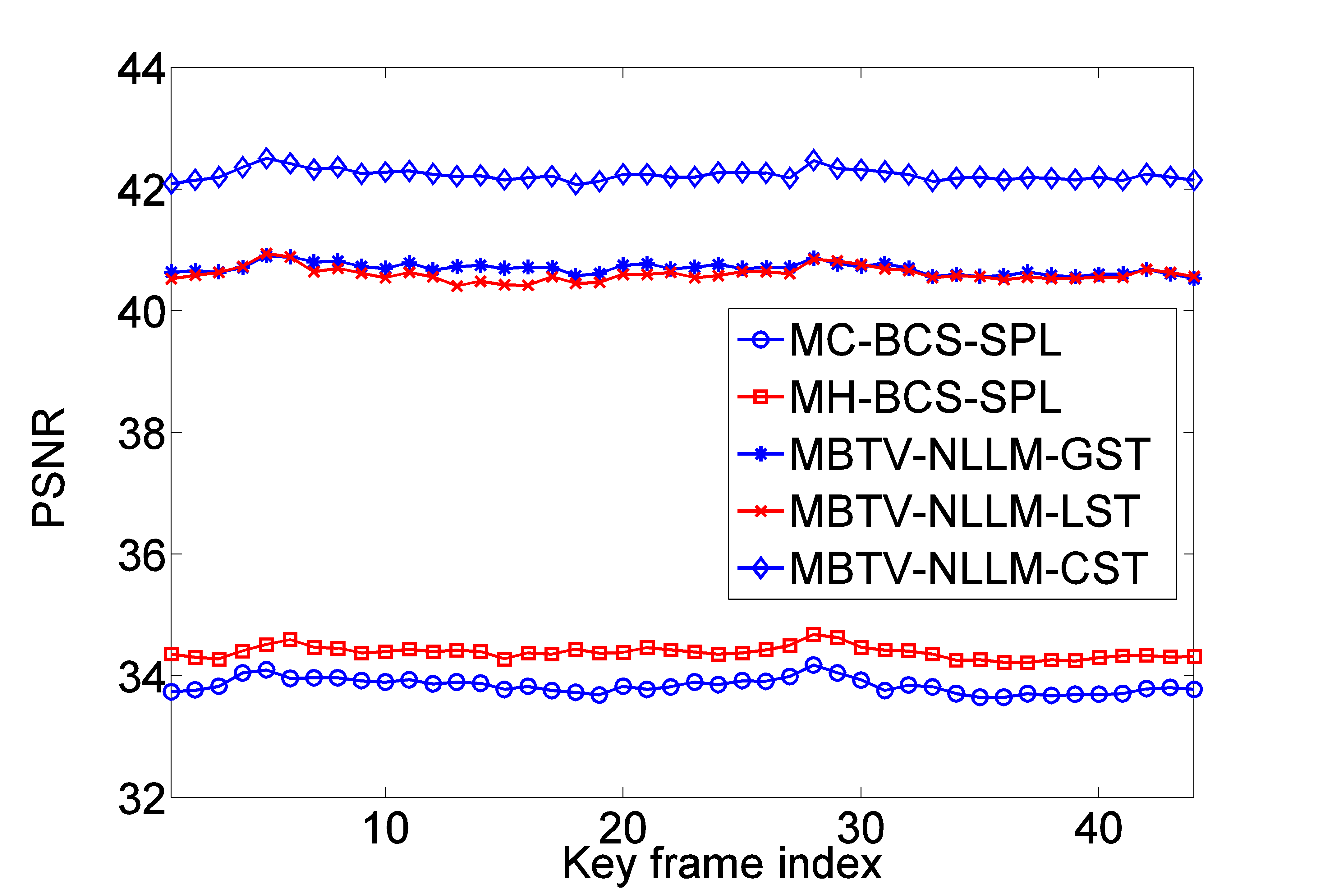}}
 		\end{tabular}}
 		\caption{Objective quality of key-frames (block-size $16 \times 16$, subrate $0.7$)} \label{fig11}
 	\end{figure*}

The effectiveness of the proposed CS recovery design is also evaluated with the first $88$ frames of three QCIF video sequences: News, Mother-daughter, and Salesman \cite{videotest}. The GOP is set to $2$. Input frames are split into non-overlapping blocks ($16 \times 16$ in size), each of which is subject to BCS by an i.i.d. random Gaussian sensing matrix. To achieve better quality, key frames are sensed with a subrate of $0.7$, while non-key frames are sensed by a subrate ranging from $0.1$ to $0.7$. Figure~\ref{fig11} shows the improvements of the reconstructed key frames of the proposed methods compared with MC-BCS-SPL and MH-BCS-SPL. All three proposed recovery schemes show far better visual quality than the previous block DCVS in \cite{mun2011residual, tramel2011video}. On average, over the three tested video sequences, MBTV-NLLM-CST shows $8.45$ dB and $7.84$ dB gains over MC-BCS-SPL and MH-BCS-SPL, respectively.

\begin{figure*}[t]
  	\centerline{%
  		\setlength\tabcolsep{1.5pt} 
  		\begin{tabular}{c c c} 
  			\subfloat[News]{\includegraphics[trim=5.5cm 0.5cm 8.5cm 4.8cm, clip=true, width=2.3in]{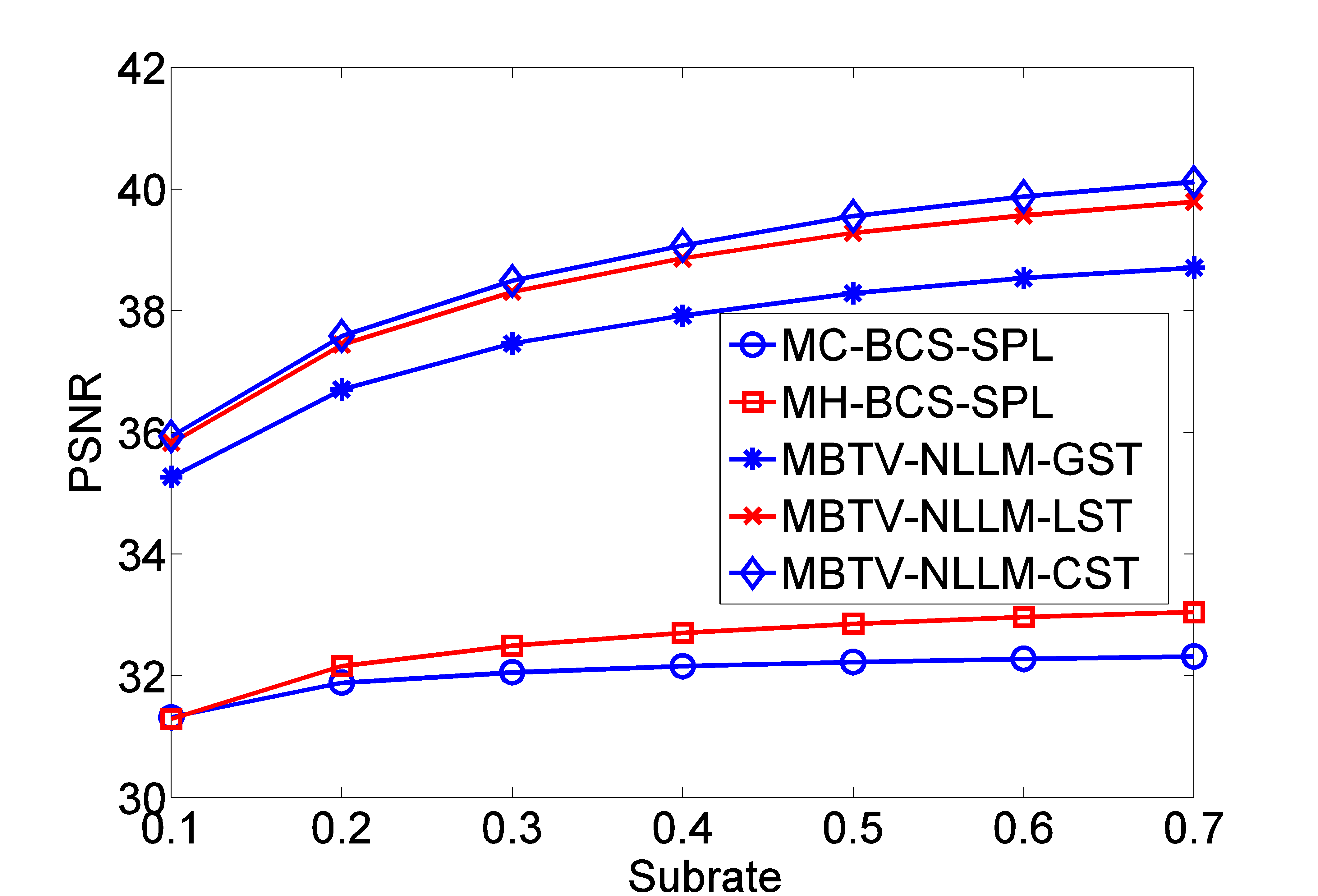}} &
  			\subfloat[Mother-daughter]{\includegraphics[trim=5.5cm 0.5cm 8.5cm 4.8cm, clip=true, width=2.3in]{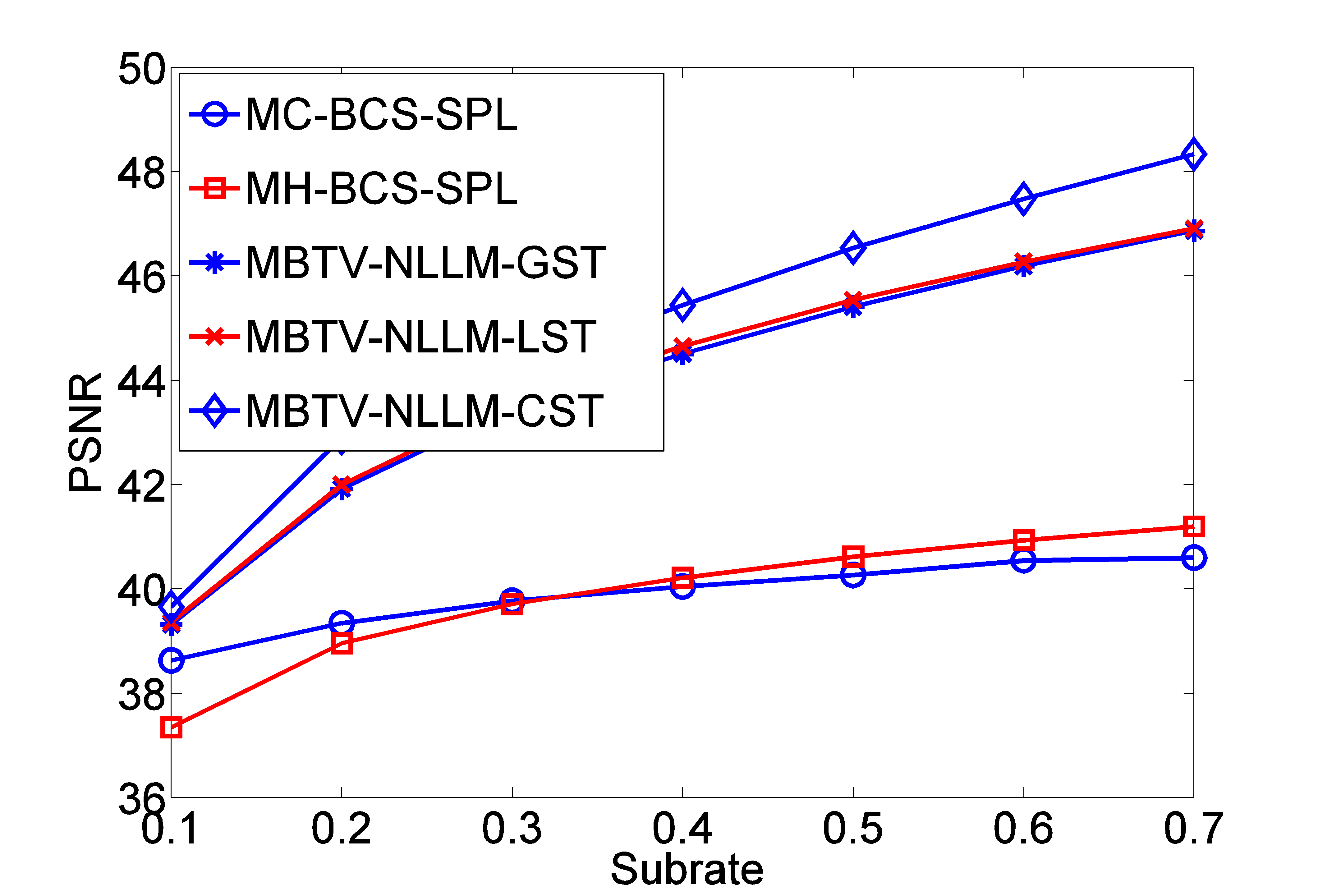}} & \subfloat[Salesman]{\includegraphics[trim=5.5cm 0.5cm 8.5cm 4.8cm, clip=true, width=2.3in]{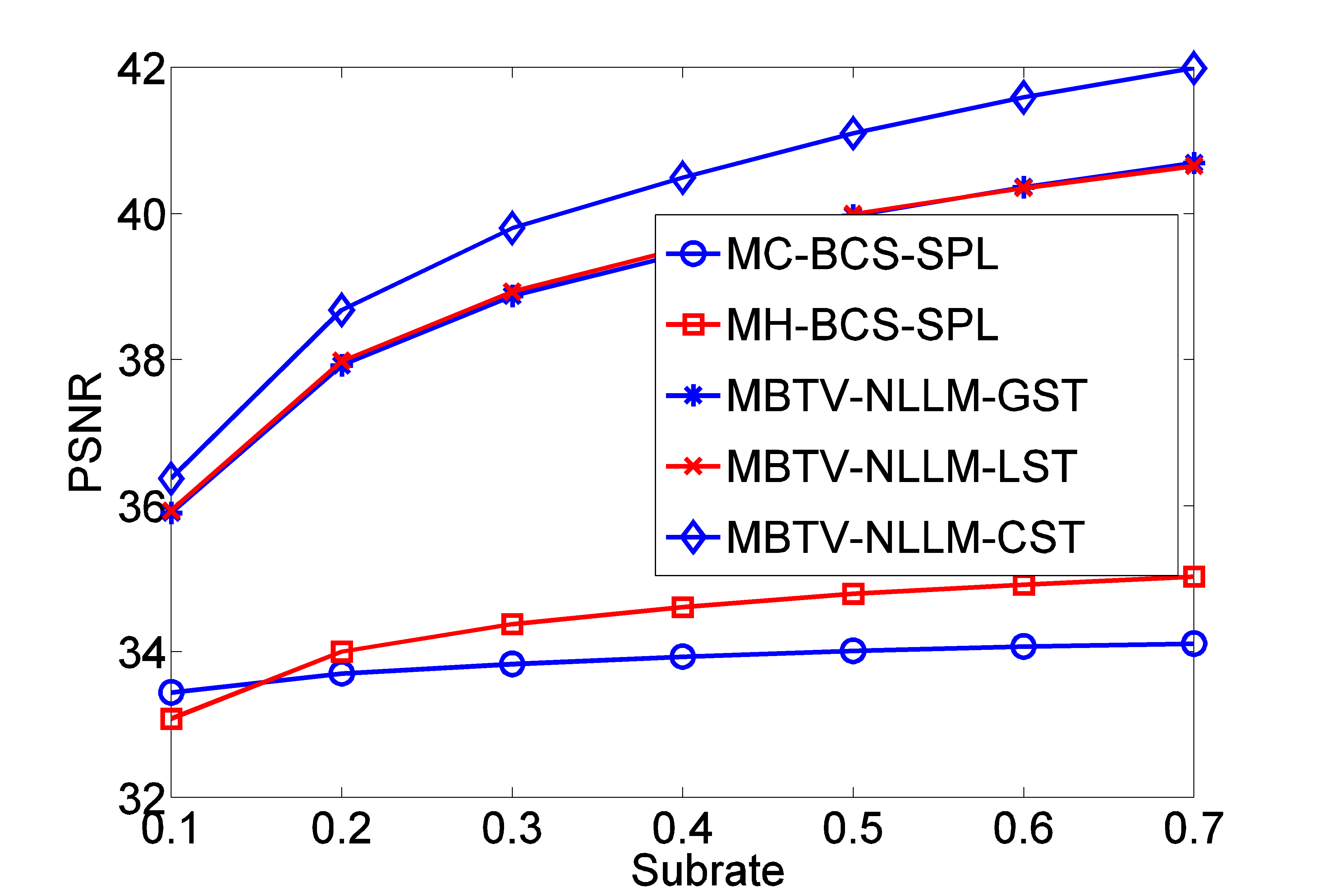}}
  		\end{tabular}}
  		\caption{Objective quality of non-key frames (block-size $16 \times 16$)} \label{fig12}
  	\end{figure*}

The improvements of non-key frames for various block DCVS are shown in Figure~\ref{fig12}. Due to the findings that $1)$ TV can preserve edge objects, $2)$ the nonlocal Lagrangian multiplier can reduce staircasing artifacts, and 3) the patch-based sparsifying transforms can enrich detail information, our proposed CS recovery schemes also produce far better PSNR values. Compared with MC-BCS-SPL, MBTV-NLLM-CST demonstrates gains between $2.31$ dB and $8.36$ dB depending on the subrate. In the best case, our recovery scheme is better by an average of $8.66$ dB compared to MC-BCS-SPL over $44$ non-key frames. 

\begin{figure*}[t]
  		\centerline{%
  			\setlength\tabcolsep{1.5pt} 
  			\begin{tabular}{c c c}
  				\subfloat[]{\includegraphics[width = 3.52cm, height = 2.88 cm]{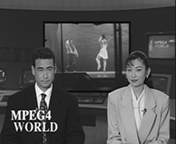}} &
  				\subfloat[]{\includegraphics[width = 3.52cm, height = 2.88 cm]{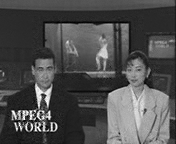}} &
  				\subfloat[]{\includegraphics[width = 3.52cm, height = 2.88 cm]{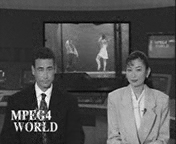}} \vspace{-0.3cm}\\
  				\subfloat[]{\includegraphics[width = 3.52cm, height = 2.88 cm]{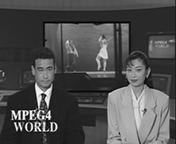}} &
  				\subfloat[]{\includegraphics[width = 3.52cm, height = 2.88 cm]{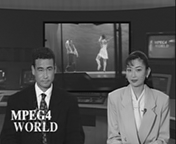}} &
  				\subfloat[]{\includegraphics[width = 3.52cm, height = 2.88 cm]{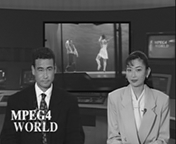}} \vspace{-0.3cm}	
  			\end{tabular}}
  			\caption{Visual quality and FSIM value of the first non-key frame of sequence News (subrate $0.1$, block size $16 \times 16$, $\ast$denotes the proposed methods: $(a)$ Ground Truth, $(b)$ MC-BCS-SPL \cite{mun2011residual} $\textrm{FSIM}=0.950$, $(c)$ MH-BCS-SPL \cite{tramel2011video} $\textrm{FSIM}=0.965$, $(d)$  MBTV-NLLM-GST$^\ast$ $\textrm{FSIM}=0.993$ , $(e)$  MBTV-NLLM-LST$^\ast$ $\textrm{FSIM}=0.993$, $(f)$  MBTV-NLLM-CST$^\ast$ $\textrm{FSIM}=0.994$
  			} \label{fig13}
  		\end{figure*}

Moreover, the visual quality of the first non-key frame of the News sequence is illustrated in Figure~\ref{fig13}. Because we utilized temporal redundancy over the frames, detailed information could be preserved for all block DCVS schemes. The high values of FSIM, even at a subrate of $0.1$, demonstrate how crucial it is to exploit the correlation of frames in compressive video sensing. However, MC-BCS-SPL \cite{mun2011residual} and MH-BCS-SPL \cite{tramel2011video} still suffer from high-frequency oscillatory artifacts. Meanwhile, the proposed schemes no longer appear to have artifacts (i.e., the FSIM values are very close to $1$).

 
  \subsection{Computational complexity}
 Excluding patch-based sparse representation, the main computational complexity of the proposed MBTV-NLLM comes from the high cost of the NLM filter. More specifically, if the search range and size of the similarity patches of the NLM filter are $[-S,S]^2$ and $(2B+1)(2B+1)$, respectively, then, for an image of size $\sqrt{N} \times \sqrt{N}$, the computational complexity of this filter is $\mathcal{O}(N(2S + 1)^2 (2B + 1)^2)$. For natural images that are $256 \times 256$ in size, with a subrate of $0.1$, MBTV-NLLM takes around $1$ min. to recover in our simulation. This is comparable to other methods that also do not use patch-based sparsifying transforms. That is, with the image Leaves at a subrate of 0.1, MBTV-NLLM needs 63 s, MH takes $40$ s, and SPLDDWT consumes $44$ s. Alternatively, TSDCT requires much more decoding time than the others, requiring about $10$ min. to recover. 
  	
Patch-based sparse representation acts as a computational bottle-neck. For a patch size of $(2B+1)(2B+1)$, search range of $[-S,S]^2$, group size of $(2B+1)^2 F$ (where $F$ is the number of similar patches in a group), and two constant values $k_1$ and $k_2$, the patch-based sparse representation using a local sparsifying transform demands a computational complexity of $\mathcal{O} \left( N ( k_1 (2B + 1)^4 F + k_2 F^3 + (2S + 1)^2 (2B + 1)^2) \right)$. In this way, MBTV-NLLM-CST is more complex due to the second stage containing the global sparsifying transform. Subsequently, for recovery of a QCIF video frame using MBTV-NLLM-CST, a key frame demands around $3$ min., and a non-key frame requires $40$ s. Therefore, complexity optimization of patch-based sparse representation is an important task for future works. Specifically, to reduce complexity, we may be able to integrate our reconstructed algorithms with a robust sensing matrix such as Gaussian regression-based \cite{han2015novel} or multi-scale-based sensing matrices \cite{fowler2011multiscale}.

 \section{Conclusion} \label{SecVI}
This paper proposed recovery schemes for BCS of still images and video that can recover pictures with high-quality performance. For compressive imaging, the modified augmented Lagrangian total variation with a multi-block gradient process and nonlocal Lagrangian multiplier are used to generate an initial recovered image. Subsequently, the patch-based sparse representation enhances the local detailed information. Our design is also easily extendible to DCVS. More specifically, key frames are reconstructed to have improved quality and used to create initial versions of non-key frames. Subsequently, non-key frames are refined by patch-based sparsifying transform-aided side information regularization. Our experimental results demonstrated the improvements made by the proposed recovery schemes compared to representative state-of-the-art algorithms for both natural images and video.
 
\section*{Acknowledgements}
This work was supported by a National Research Foundation of Korea (NRF) grant funded by the Korean government (MSIP) (No.$2011$-$001$-$7578$), by the MSIP G-ITRC support program (IITP-2016-R6812-16-0001) supervised by the IITP, and by the ERC via Grant EU FP 7 - ERC Consolidator Grant 615216 LifeInverse.

\section*{References}

\bibliography{mybibfile}

\end{document}